\documentclass{article}

\usepackage{graphicx,amscd,amsmath,amssymb,verbatim}
\usepackage[dvips]{hyperref}
\usepackage[TS1,OT1,T1]{fontenc}

\begin{document}

\title{Modelling the Probability Density of Markov Sources\footnote{Submitted to Neural Computation on 2 February 1998. Manuscript no. 1729. It was not accepted for
publication, but it underpins several subsequently published papers.}}
\author{Stephen Luttrell}
\maketitle

\noindent {\bfseries Abstract:} This paper introduces an objective function that seeks to minimise the average
total number of bits required to encode the joint state of all of the layers of a Markov source. This type of
encoder may be applied to the problem of optimising the bottom-up (recognition model) and top-down (generative
model) connections in a multilayer neural network, and it unifies several previous results on the optimisation
of multilayer neural networks.

\section{Introduction}

There is currently a great deal of interest in modelling probability density
functions (PDF). This research is motivated by the fact that the joint PDF of
a set of variables can be used to deduce any conditional PDF which involves
these variables alone, which thus allows all inference problems in the space
of these variables to be addressed quantitatively. The only limitation of this
approach to solving inference problems is that a model of the PDF is used,
rather than the actual PDF itself, which can lead to inaccurate inferences.
The objective function for optimising a PDF model is usually to maximise the
log-likelihood that it could generate the training set: i.e.
maximise\ $\left\langle \log\left(  \text{model probability}\right)
\right\rangle _{\text{training set}}$.

In this paper the problem of modelling the PDF of a Markov source will be
studied. In the language of neural networks, this type of source can be viewed
as a layered network, in which the state of each layer directly influences the
states of only the layers immediately above and below it. The optimal PDF
model then approximates the joint PDF of the states of all of the layers of
the network, or at least some subset of the layers of the network, which is a
generalisation of what is conventionally done in neural network PDF models.

Markov source density modelling is interesting because it unifies a number of
existing neural network techniques into a single framework, and may also be
viewed as a density modelling perspective on the results reported in
\cite{Luttrell1994a}. For instance, an approximation to the standard Kohonen
topographic mapping neural network \cite{Kohonen1989} emerges from density
modelling the joint PDF of the input and output layers of a 3-layer network,
and generalisations of the Kohonen network also emerge naturally from this framework.

In section \ref{Sect:CodingTheory} the relevant parts of the Shannon theory of
information are summarised \cite{Shannon1948}, and the application to coding
various types of source is derived \cite{Rissanen1978}; in particular, Markov
sources are discussed, because they are the key to the approach that is
presented in this paper. In section \ref{Sect:ApplicationNN} the application
of Markov source coding to unsupervised neural networks is discussed in
detail, including the Kohonen network \cite{Kohonen1989}. In section
\ref{Sect:ACE} (and appendix \ref{Appendix:ACE})\ hierarchical encoding using
an adaptive cluster expansion (ACE) is discussed \cite{Luttrell1989c}, and in
section \ref{Sect:PMD} (and appendix \ref{Appendix:PMD})\ factorial encoding
using a partitioned mixture distribution (PMD) are discussed
\cite{Luttrell1993}. Finally in appendix \ref{Appendix:Helmholtz} density
modelling of Markov sources is compared with standard density modelling using
a Helmholtz machine \cite{Dayan1995}.

\section{Coding Theory}

\label{Sect:CodingTheory}In section \ref{Sect:InformationTheory} the basic
ideas of information theory are outlined (this discussion is inspired by the
reasoning presented in \cite{Shannon1948}), and in section
\ref{Sect:SourceCoding} the process of using a model to code a source
described in detail. In section \ref{Sect:MarkovSourceCoding} this is extended
to the case of a Markov source. In section \ref{Sect:AlternativeViewpoints}
the relationship between conventional density models and Markov density models
is discussed.

See \cite{Shannon1948} for a lucid introduction to information theory, and see
\cite{Rissanen1978} for a discussion of the number of bits required to encode
a source using a model.

\subsection{Information Theory}

\label{Sect:InformationTheory}A source of symbols (drawn from an alphabet of
$M$ distinct symbols) is modelled as a vector of probabilities denoted as
$\mathbf{P}$%
\begin{equation}
\mathbf{P\equiv}\left(  P_{1},P_{2},\cdots,P_{M}\right)
\end{equation}
which describes the relative frequency with which each symbol is drawn
independently from the source $\mathbf{P}$. A trivial example is an unbiassed
die, which has $M=6$ and $P_{i}=\frac{1}{6}$ for $i=1,2,\cdots,6$.

The ordered sequence of symbols drawn independently from a source may be
partitioned into subsequences of $N$ symbols, and each such subsequence will
be called a message. If $N$ is very large, then a message is called ``likely''
if the relative frequency of occurrence of its symbols approximates
$\mathbf{P}$, otherwise it is called ``unlikely''. As $N\rightarrow\infty$ the
set of likely messages is very sharply defined, in the sense that the
proportion of all messages that lie in the transition region between being
likely and being unlikely becomes vanishingly small. Thus there is a set of
likely messages all with equal probability of occurring (because each likely
message has the same relative frequency of occurrence of each of the $M$
possible symbols), and a set of unlikely messages (i.e. all the messages that
are not likely messages) that have essentially zero probability of occurring.
It is this separation of messages into a likely set (all with equal
probability)\ and an unlikely set (all with zero probability) that underlies
information theory, as discussed in \cite{Shannon1948}.

A likely message from $\mathbf{P}$ will be called a likely $\mathbf{P}%
$-message. As $N\longrightarrow\infty$ the number of times $n_{i}$ that each
symbol $i$ occurs in a $\mathbf{P}$-message of length $N$ is $n_{i}=NP_{i}$,
where $\sum_{i=1}^{M}P_{i}=1$ guarantees that the normalisation condition
$\sum_{i=1}^{M}n_{i\,}=N$ is satisfied. The logarithm of the number of
different likely $\mathbf{P}$-messages is given by (using Stirling's
approximation $\log x!\approx x\log x-x$ when $x$ is large)
\begin{equation}
\log\left(  \frac{N!}{n_{1}!n_{2}!\cdots n_{M}!}\right)  \approx-N\sum
_{i=1}^{M}P_{i}\log P_{i}\label{Eq:LikelyMessage}%
\end{equation}

Now define the entropy $H\left(  \mathbf{P}\right)  $ of source $\mathbf{P}$
as the logarithm of the number of different likely $\mathbf{P}$-messages
(measured per message symbol)\textbf{:}
\begin{equation}
H\left(  \mathbf{P}\right)  \equiv-\sum_{i=1}^{M}P_{i}\log P_{i}%
\geq0\label{Eq:Entropy}%
\end{equation}
Thus $H\left(  \mathbf{P}\right)  $ is the number of bits per symbol (on
average) required to encode the source (assuming a perfect encoder), because
the only messages that the source has a finite probability of producing are
the likely $\mathbf{P}$-messages that are enumerated in equation
\ref{Eq:LikelyMessage}.

It is usually very difficult to encode the source $\mathbf{P}$ using $H\left(
\mathbf{P}\right)  $ bits per symbol on average. This is because although the
boundary between the set of likely $\mathbf{P}$-messages and the set of
unlikely $\mathbf{P}$-messages is sharply defined in principle, in practice it
is very hard to model mathematically. If this boundary is not precisely
defined, then it is impossible to compute the value of $H\left(
\mathbf{P}\right)  $ accurately. In order to ensure that all of the likely
$\mathbf{P}$-messages are accounted for, it is necessary for the mathematical
model of the boundary to lie outside the true boundary, which thus
overestimates the value of $H\left(  \mathbf{P}\right)  $. This demonstrates
that $H\left(  \mathbf{P}\right)  $ is in fact a lower bound on the true
number of bits per symbol that must be used to encode the source $\mathbf{P}$.

\subsection{Source Coding}

\label{Sect:SourceCoding}The mathematical model (or, simply, the model) of the
boundary between the set of likely $\mathbf{P}$-messages and the set of
unlikely $\mathbf{P}$-messages may be derived from a another vector of
probabilities, denoted as $\mathbf{Q}$, whose $M$ elements model the
probability of each symbol drawn from an alphabet of $M$ distinct symbols. If
$\mathbf{Q=P}$ then the boundary is modelled perfectly, and hence in principle
the lower bound $H\left(  \mathbf{P}\right)  $ on the number of bits per
symbol may be attained, although even this is usually difficult to realise
constructively in practice. In practical situations $\mathbf{Q\neq P} $ is
invariably the case, so the problem of coding a source with an inaccurate
model cannot be avoided.

Since the only $\mathbf{P}$-messages that can occur are the likely
$\mathbf{P}$-messages (which all occur with equal probability), the number of
bits required when using $\mathbf{Q}$ to encode $\mathbf{P}$ is (minus) the
logarithm of the probability $\Pi_{N}\left(  \mathbf{P},\mathbf{Q}\right)  $
that a $\mathbf{Q}$-message is one of the likely $\mathbf{P}$-messages.
$\Pi_{N}\left(  \mathbf{P},\mathbf{Q}\right)  $ is given by
\begin{align}
\Pi_{N}\left(  \mathbf{P},\mathbf{Q}\right)   & =\log\left(  \frac{N!}%
{n_{1}!n_{2}!\cdots n_{M}!}\,Q_{1}^{n_{1}}Q_{2}^{n_{2}}\cdots Q_{M}^{n_{M}%
}\right) \nonumber\\
& \approx-N\sum_{i=1}^{M}P_{i}\log\frac{P_{i}}{Q_{i}}\leq0
\end{align}
which is negative because the model $\mathbf{Q}$ generates likely $\mathbf{P}
$-messages with less than unit probability. The model $\mathbf{Q}$ must be
used to generate enough $\mathbf{Q}$-messages to ensure that all of the likely
$\mathbf{P}$-messages are reproduced, which requires the basic $H\left(
\mathbf{P}\right)  $ bits per symbol (that would be required if $\mathbf{Q=P}%
$), plus some extra bits to compensate for the less than $100\%$ efficiency
with which $\mathbf{Q}$ generates likely $\mathbf{P}$-messages (because
$\mathbf{Q\neq P}$). The number of extra bits per symbol is the relative
entropy $G\left(  \mathbf{P},\mathbf{Q}\right)  $%
\begin{equation}
G\left(  \mathbf{P},\mathbf{Q}\right)  \equiv\sum_{i=1}^{M}P_{i}\log
\frac{P_{i}}{Q_{i}}\geq0\label{Eq:RelativeEntropy}%
\end{equation}
which is $-\frac{\Pi_{N}\left(  \mathbf{P},\mathbf{Q}\right)  }{N}$, or minus
the logarithm of the probability per symbol that a $\mathbf{Q}$-message is a
likely $\mathbf{P}$-message. Thus $\mathbf{Q}$ is used to generate exactly the
number of extra $\mathbf{Q}$-messages required to compensate for the fact that
the probability that each $\mathbf{Q}$-message is a likely $\mathbf{P}%
$-message is less than unity (i.e. $\Pi_{N}\left(  \mathbf{P},\mathbf{Q}%
\right)  \leq0$). $G\left(  \mathbf{P},\mathbf{Q}\right)  $ (i.e. relative
entropy) is the amount by which the number of bits per symbol exceeds the
lower bound $H\left(  \mathbf{P}\right)  $ (i.e. source entropy). For
completeness, also define the total number of bits per symbol $H\left(
\mathbf{P}\right)  +G\left(  \mathbf{P},\mathbf{Q}\right)  $ as $L\left(
\mathbf{P},\mathbf{Q}\right)  $, which is given by
\begin{align}
L\left(  \mathbf{P},\mathbf{Q}\right)   & \equiv H\left(  \mathbf{P}\right)
+G\left(  \mathbf{P},\mathbf{Q}\right) \nonumber\\
& =-\sum_{i=1}^{M}P_{i}\log Q_{i}\geq0
\end{align}

The expression for $G\left(  \mathbf{P},\mathbf{Q}\right)  $ provides a means
of optimising the model $\mathbf{Q}$. If the optimisation criterion is that
the average number of bits per symbol required when using $\mathbf{Q}$ to
encode $\mathbf{P}$ should be minimised, then the optimum model $\mathbf{Q}%
_{opt}$ should minimise the objective function $G\left(  \mathbf{P}%
,\mathbf{Q}\right)  $ with respect to $\mathbf{Q}$, thus
\begin{equation}
\mathbf{Q}_{opt}=
\begin{array}
[c]{c}%
\arg\min\\
\mathbf{Q}%
\end{array}
G\left(  \mathbf{P},\mathbf{Q}\right)
\end{equation}
This criterion for optimising a model does not include the number of bits
required to specify the model itself, such as is used in the minimum
description length approach \cite{Rissanen1978}, although the objective
function could be extended to include such additional contributions.

$G\left(  \mathbf{P},\mathbf{Q}\right)  $ is frequently used as an objective
function in density modelling, where the source $\mathbf{P}$ is the vector of
observed symbol frequencies. Since $\mathbf{Q}_{opt}$ must, in some sense, be
close to $\mathbf{P}$, this affords a practical way of ensuring that the
optimum model probabilities $\mathbf{Q}_{opt}$ are similar to the source
symbol frequencies $\mathbf{P}$, which is the goal of density modelling.

\subsection{Markov Source Coding}

\label{Sect:MarkovSourceCoding}The above scheme for using a model $\mathbf{Q}
$ to encode symbols derived from a source $\mathbf{P}$ may be extended to the
case where the source and the model are $L$-layer first order Markov chains.
The word ``layer'' is used in anticipation of the connection with multilayer
neural networks that will be discussed in section \ref{Sect:ApplicationNN}.
Thus split up both of $\mathbf{P\,}$and $\mathbf{Q}$ into their constituent
transition probabilities
\begin{align}
\mathbf{P}  & =\left(  \mathbf{P}^{0},\,\mathbf{P}^{1|0},\cdots\,,\mathbf{P}%
^{L-1|L-2},\mathbf{P}^{L|L-1}\right) \nonumber\\
& =\left(  \mathbf{P}^{0|1},\mathbf{P}^{1|2},\,\cdots,\mathbf{P}%
^{L-1|L},\mathbf{P}^{L}\right) \nonumber\\
\mathbf{Q}  & =\left(  \mathbf{Q}^{0},\mathbf{Q}^{1|0},\,\cdots,\mathbf{Q}%
^{L-1|L-2},\mathbf{Q}^{L|L-1}\right) \nonumber\\
& =\left(  \mathbf{Q}^{0|1},\mathbf{Q}^{1|2},\cdots,\mathbf{Q}^{L-1|L}%
,\mathbf{Q}^{L}\right) \label{Eq:MarkovChain}%
\end{align}
These two ways of decomposing $\mathbf{P}$ (and $\mathbf{Q}$) are equivalent,
because a forward pass through a Markov chain may be converted into a backward
pass through a different Markov chain, whose transition probabilities are
uniquely determined by applying Bayes' theorem to the original Markov chain.
$\mathbf{P}^{k|l}$ (and $\mathbf{Q}^{k|l}$)\ is the matrix of transition
probabilities from layer $l$ to layer $k$ of the Markov chain of the source
(and model), $\mathbf{P}^{0}$ ($\mathbf{Q}^{0}$) is the vector of marginal
probabilities in layer $0$, $\mathbf{P}^{L}$ (and $\mathbf{Q}^{L}$) is the
vector of marginal probabilities in layer $L$. This may be written out in
detail as%

\begin{align}
P_{i_{0}}^{0}  & =\text{true probability that layer }0\text{ has state }%
i_{0}\nonumber\\
P_{i_{L}}^{L}  & =\text{true probability that layer }L\text{ has state }%
i_{L}\nonumber\\
P_{i_{l+1},i_{l}}^{l+1|l}  & =\text{true probability that layer }l+1\text{ has
state }i_{l+1}\nonumber\\
& \text{given that layer }l\text{ has state }i_{l}\nonumber\\
P_{i_{l},i_{l+1}}^{l|l+1}  & =\text{true probability that layer }l\text{ has
state }i_{l}\nonumber\\
& \text{given that layer }l+1\text{ has state }i_{l+1}%
\end{align}%
\begin{align}
Q_{i_{0}}^{0}  & =\text{model probability that layer }0\text{ has state }%
i_{0}\nonumber\\
Q_{i_{L}}^{L}  & =\text{model probability that layer }L\text{ has state }%
i_{L}\nonumber\\
Q_{i_{l+1},i_{l}}^{l+1|l}  & =\text{model probability that layer }l+1\text{
has state }i_{l+1}\nonumber\\
& \text{given that layer }l\text{ has state }i_{l}\nonumber\\
Q_{i_{l},i_{l+1}}^{l|l+1}  & =\text{model probability that layer }l\text{ has
state }i_{l}\nonumber\\
& \text{given that layer }l+1\text{ has state }i_{l+1}%
\end{align}

The number of extra bits per symbol $G\left(  \mathbf{P},\mathbf{Q}\right)  $
(see equation \ref{Eq:RelativeEntropy})\ required to encode each symbol from
the source $\mathbf{P}$ using the model $\mathbf{Q}$ may then be written as
\begin{align}
G\left(  \mathbf{P},\mathbf{Q}\right)   & =\sum_{i_{0}=1}^{M_{0}}\cdots
\sum_{i_{L}=1}^{M_{L}}P_{i_{0},i_{1}}^{0|1}\,P_{i_{1},i_{2}}^{1|2}%
\,\cdots\,P_{i_{L-1},i_{L}}^{L-1|L}\,P_{i_{L}}^{L}\nonumber\\
& \times\log\left(  \frac{P_{i_{0},i_{1}}^{0|1}\,P_{i_{1},i_{2}}^{1|2}%
\,\cdots\,P_{i_{L-1},i_{L}}^{L-1|L}\,P_{i_{L}}^{L}}{Q_{i_{0},i_{1}}%
^{0|1}\,Q_{i_{1},i_{2}}^{1|2}\,\cdots\,Q_{i_{L-1},i_{L}}^{L-1|L}\,Q_{i_{L}%
}^{L}}\right) \nonumber\\
& =\sum_{l=0}^{L-1}\sum_{i_{l+1}=1}^{M_{l+1}}P_{i_{l+1}}^{l+1}G_{i_{l+1}%
}\left(  \mathbf{P}^{l|l+1},\mathbf{Q}^{l|l+1}\right)  +G\left(
\mathbf{P}^{L},\mathbf{Q}^{L}\right)
\end{align}
where the flow of influence in both $\mathbf{P}$ and $\mathbf{Q}$ is from
layer $0$ to layer $L$. The suffix $i_{l+1}$ that appears on the $G_{i_{l+1}%
}\left(  \mathbf{P}^{l|l+1},\mathbf{Q}^{l|l+1}\right)  $ indicates that the
state of layer $l+1$ is fixed during the evaluation of $G_{i_{l+1}}\left(
\mathbf{P}^{l|l+1},\mathbf{Q}^{l|l+1}\right)  $ (i.e. it is the relative
entropy of layer $l$, given that the state of layer $l+1$ is known).$\,\,$%
Similarly, the total number of bits per symbol required to encode each symbol
from the source $\mathbf{P}$ using the model $\mathbf{Q}$ is $L\left(
\mathbf{P},\mathbf{Q}\right)  $ (i.e. $H\left(  \mathbf{P}\right)  +G\left(
\mathbf{P},\mathbf{Q}\right)  $), which is given by
\begin{equation}
L\left(  \mathbf{P},\mathbf{Q}\right)  =\sum_{l=0}^{L-1}\sum_{i_{l+1}%
=1}^{M_{l+1}}P_{i_{l+1}}^{l+1}L_{i_{l+1}}\left(  \mathbf{P}^{l|l+1}%
,\mathbf{Q}^{l|l+1}\right)  +L\left(  \mathbf{P}^{L},\mathbf{Q}^{L}\right)
\label{Eq:NegativeLogLikelihood}%
\end{equation}

This result has a very natural interpretation. Both the source $\mathbf{P}$
and the model $\mathbf{Q}$ are Markov chains, and corresponding parts of the
model are matched up with corresponding parts of the source. First of all, the
number of bits required to encode the $L^{th}$ layer of the source is
$L\left(  \mathbf{P}^{L},\mathbf{Q}^{L}\right)  $. Having done that, the
number of bits required to encode the $L-1^{th}$ layer of the source, given
that the state of the $L^{th}$ layer is already known, is $L\left(
\mathbf{P}^{L-1|L},\mathbf{Q}^{L-1|L}\right)  $, which must then be averaged
over the alternative possible states of the $L^{th}$ layer to yield
$\sum_{i_{L}=1}^{M_{L}}P_{i_{L}}^{L}L\left(  \mathbf{P}^{L-1|L},\mathbf{Q}%
^{L-1|L}\right)  $. This process is then repeated to encode the $L-2^{th}$
layer of the source, given that the state of the $L-1^{th}$ layer is already
known, and so on back to layer $0$. This yields precisely the expression for
$L\left(  \mathbf{P},\mathbf{Q}\right)  $ given above.

Bayes' theorem (in the form $P_{i_{l+1}}^{l+1}\,P_{i_{l},i_{l+1}}%
^{l|l+1}=P_{i_{l}}^{l}\,P_{i_{l+1},i_{l}}^{l+1|l}$) may be used to rewrite the
expression for $L\left(  \mathbf{P},\mathbf{Q}\right)  $ so that the flow of
influence in $\mathbf{P}$ and $\mathbf{Q}$ runs in opposite directions. Thus
\begin{equation}
L\left(  \mathbf{P},\mathbf{Q}\right)  =\sum_{l=0}^{L-1}\sum_{i_{l}=1}^{M_{l}%
}P_{i_{l}}^{l}K_{i_{l}}\left(  \mathbf{P}^{l+1|l},\mathbf{Q}^{l|l+1}\right)
+L\left(  \mathbf{P}^{L},\mathbf{Q}^{L}\right)
\label{Eq:ObjectiveMarkovSource}%
\end{equation}
where $K_{i_{l}}\left(  \mathbf{P}^{l+1|l},\mathbf{Q}^{l|l+1}\right)  $ is
defined as
\begin{equation}
K_{i_{l}}\left(  \mathbf{P}^{l+1|l},\mathbf{Q}^{l|l+1}\right)  \equiv
-\sum_{i_{l+1}=1}^{M_{l+1}}\,P_{i_{l+1},i_{l}}^{l+1|l}\,\log Q_{i_{l},i_{l+1}%
}^{l|l+1}%
\end{equation}
The expression for $L\left(  \mathbf{P},\mathbf{Q}\right)  $ in equation
\ref{Eq:ObjectiveMarkovSource} has an analogous interpretation to that in
equation \ref{Eq:NegativeLogLikelihood}.

Other types of Markov chain may also be considered, such as ones in which some
of the layers are not included in the calculation of the number of bits
required to encode the source. One such example is discussed in section
\ref{Sect:Kohonen}.

\subsection{Alternative Viewpoints}

\label{Sect:AlternativeViewpoints}The relationship between conventional
density models and Markov density models can be stated from the point of view
of a conventional density modeller. The goal is to build a density model
$\mathbf{Q}^{0}$ of the source $\mathbf{P}^{0}$, such that the number of bits
per symbol $L\left(  \mathbf{P}^{0},\mathbf{Q}^{0}\right)  $ required to
encode $\mathbf{P}^{0}$ is minimised. However, if the source $\mathbf{P}^{0}$
is transformed through $L$ layers of a network to produce a transformed source
$\mathbf{P}^{L}$, then $L\left(  \mathbf{P}^{0},\mathbf{Q}^{0}\right)  \leq
L\left(  \mathbf{P},\mathbf{Q}\right)  $ where $L\left(  \mathbf{P}%
,\mathbf{Q}\right)  $ is given in equation \ref{Eq:NegativeLogLikelihood}%
,$\,$which is the sum of the number of bits per symbol $L\left(
\mathbf{P}^{L},\mathbf{Q}^{L}\right)  $ required to encode $\mathbf{P}^{L}$,
plus (for $l=0,1,\cdots,L-1$) the number of bits per symbol $\sum_{i_{l+1}%
=1}^{M_{l+1}}P_{i_{l+1}}^{l+1}L_{i_{l+1}}\left(  \mathbf{P}^{l|l+1}%
,\mathbf{Q}^{l|l+1}\right)  \,$required to encode $\mathbf{P}^{l|l+1}$.

Thus the problem of encoding the source $\mathbf{P}^{0}$ can be split into
three steps:\ transform the source from $\mathbf{P}^{0}$ to $\mathbf{P}^{L}$,
encode the transformed source $\mathbf{P}^{L}$, and encode all of the
transformations $\mathbf{P}^{l|l+1}$ (for $l=0,1,\cdots,L-1$) to allow the
original source to be reconstructed from the transformed source. The total
number of bits $L\left(  \mathbf{P},\mathbf{Q}\right)  $ required to encode
$\mathbf{P}^{L}$ and $\mathbf{P}^{l|l+1}$ (for $l=0,1,\cdots,L-1$) is then an
upper bound on the total number of bits $L\left(  \mathbf{P}^{0}%
,\mathbf{Q}^{0}\right)  $ required to encode $\mathbf{P}^{0}$. In this
picture, a Markov chain is used to connect the original source $\mathbf{P}%
^{0}$ to the transformed source $\mathbf{P}^{L}$, so the Markov chain relates
one conventional density modelling problem (i.e. optimising $\mathbf{Q}^{0}$)
to another (i.e. optimising $\mathbf{Q}^{L}$).

The above description of the relationship between conventional density models
and Markov density models was presented from the point of view of a
conventional density modeller, who asserts that the goal is to build an
optimum (i.e. minimum number of bits per symbol) density model $\mathbf{Q}%
^{0}$ of the source $\mathbf{P}^{0}$. From this point of view, the Markov
chain is merely a means of transforming the problem from modelling the source
$\mathbf{P}^{0}$ to modelling the transformed source $\mathbf{P}^{L}$. That
this transformation process is imperfect is reflected in the fact that more
bits per symbol are required to encode the transformed source $\mathbf{P}^{L}$
(plus the state of the Markov chain that generates it) than the original
source $\mathbf{P}^{0}$. A conventional density modeller might reasonably ask
what is the point of using Markov density models, if they give only an upper
bound on the number of bits per symbol for encoding the original source
$\mathbf{P}^{0}$?

However, it is not at all clear that the conventional density modeller is
using the correct objective function in the first place. Why should the number
of bits per symbol for encoding the original source $\mathbf{P}^{0}$ be
especially important?\ It is as if the world has been separated into an
external world (i.e. $\mathbf{P}^{0}$) and an internal world (i.e. the
$\mathbf{P}^{l+1|l}$ for $l=0,1,\cdots,L-1$), and a special status is accorded
to the external world, which deems that it is important to model its density
$\mathbf{P}^{0}$ accurately, at the expense of modelling the $\mathbf{P}%
^{l+1|l}$ accurately. In the Markov density modelling approach, this
artificial boundary between external and internal worlds is removed, because
the Markov chain models the joint density $\left(  \mathbf{P}^{0}%
,\mathbf{P}^{1|0},\cdots,\mathbf{P}^{L|L-1}\right)  $, where $\mathbf{P}^{0}$
and the $\mathbf{P}^{l+1|l}$ are all accorded equal status. This even-handed
approach is much more natural than one in which a particular part of the
source (i.e. the external source)\ is accorded a special status.%

\sloppypar
In the language of multilayer neural networks, the vector $\left(
\mathbf{P}^{0},\mathbf{P}^{0|1},\cdots,\mathbf{P}^{L|L-1}\right)  $ is the
source which comprises the bottom-up transformations (or recognition models)
which generate the states of the internal layers of the network, and the
vector $\left(  \mathbf{Q}^{0|1},\mathbf{Q}^{1|2},\cdots,\mathbf{Q}%
^{L}\right)  $ is the model of the source which comprises the top-down
transformations (or generative models). Thus the network is self-referential,
because it forms a model of a source that includes its own internal states.
This self-referential behaviour is present in both the conventional density
modelling and in the Markov density modelling approaches, but whereas in the
former case it is not optimised in an even-handed fashion, in the latter case
it is optimised in an even-handed fashion.

\section{Application To Unsupervised Neural Networks}

\label{Sect:ApplicationNN}In section \ref{Sect:SourceModelNN} the theory of
Markov source coding (that was presented in section
\ref{Sect:MarkovSourceCoding}) is applied to a multilayer neural network. In
section \ref{Sect:2LayerFMC} this approach is applied to a 2-layer neural
network to obtain a soft vector quantiser (VQ), which is generalised to a
multilayer neural network in section \ref{Sect:CoupledFMC} to obtain a network
of coupled soft VQs. In section \ref{Sect:Kohonen} it is shown how an
approximation to Kohonen's topographic mapping network can be derived from the
theory of Markov source coding. Finally, some additional results are briefly
dicussed in section \ref{Sect:AdditionalResults}.

\subsection{Source Model of a Layered Network}

\label{Sect:SourceModelNN}In this section the optimisation of the joint PDF of
the states of all of the layers of an $\left(  L+1\right)  $-layer\ encoder of
the type that was discussed in section \ref{Sect:MarkovSourceCoding} will be
considered. It turns out that this leads to new insights into the optimisation
of a multilayer unsupervised neural networks.

The Markov chain source $\mathbf{P}=\left(  \mathbf{P}^{0},\mathbf{P}%
^{1|0},\cdots,\mathbf{P}^{L-1|L-2},\mathbf{P}^{L|L-1}\right)  $ (or,
equivalently, $\mathbf{P}=\left(  \mathbf{P}^{0|1},\mathbf{P}^{1|2}%
,\cdots,\mathbf{P}^{L-1|L},\mathbf{P}^{L}\right)  $)\ may be used to describe
the true behaviour (i.e. not merely a model) of a layered neural network as
follows. $\mathbf{P}^{0}$ is an external source, and $\left(  \mathbf{P}%
^{1|0},\cdots,\mathbf{P}^{L-1|L-2},\mathbf{P}^{L|L-1}\right)  $ is an internal
source, where external/internal describes whether the source is outside/inside
the layered network, respectively. $\mathbf{P}^{l+1|l}$ is not part of the
source itself (i.e. the external source), rather it is a transition matrix
that describes the way in which the state of layer $l$ of the neural network
influences the state of layer $l+1$. There is an analogous interpretation of
$\mathbf{P}^{L}$ and the $\mathbf{P}^{l|l+1}$.

The Markov chain model $\mathbf{Q}=\left(  \mathbf{Q}^{0},\mathbf{Q}%
^{1|0},\cdots,\mathbf{Q}^{L-1|L-2},\mathbf{Q}^{L|L-1}\right)  $ (or,
equivalently, $\mathbf{Q}=\left(  \mathbf{Q}^{0|1},\mathbf{Q}^{1|2}%
,\cdots,\mathbf{Q}^{L-1|L},\mathbf{Q}^{L}\right)  $) may then be used as a
model (i.e. not actually the true behaviour) of a layered neural network.
$\mathbf{Q}$ has an analogous interpretation to $\mathbf{P}$, except that it
is a model of the source, rather than the true behaviour of the source.%

\sloppypar
It turns out to be useful for the true Markov behaviour (i.e. $\mathbf{P}$)
and the model Markov behaviour (i.e. $\mathbf{Q}$) to run in opposite
directions through the Markov chain. Thus $\mathbf{P=}\left(  \mathbf{P}%
^{0},\mathbf{P}^{1|0},\cdots,\mathbf{P}^{L-1|L-2},\mathbf{P}^{L|L-1}\right)  $
(flow of influence from layer $0$ to layer $L$ of the Markov chain) and
$\mathbf{Q}=\left(  \mathbf{Q}^{0|1},\mathbf{Q}^{1|2},\cdots,\mathbf{Q}%
^{L-1|L},\mathbf{Q}^{L}\right)  $ (flow of influence from layer $L$ to layer
$0$ of the Markov chain). In the conventional language of neural networks,
$\mathbf{P}$ is a ``recognition model'' and $\mathbf{Q}$ is a ``generative
model''. Note that the use of the word ``model'' in the terminology
``recognition model'' is strictly speaking not accurate in this context,
because $\mathbf{P}$ is a source, not a model. However, terminology depends on
one's viewpoint. In Markov chain density modelling $\mathbf{P}$ is a source
when viewed from the point of view of the model $\mathbf{Q}$. Whereas, in
conventional density modelling $\mathbf{P}^{0}$ is a source when viewed from
the point of model $\mathbf{Q}^{0}$, in which case $\left(  \mathbf{P}%
^{1|0},\cdots,\mathbf{P}^{L-1|L-2},\mathbf{P}^{L|L-1}\right)  $ is a
recognition model and $\left(  \mathbf{Q}^{0|1},\mathbf{Q}^{1|2}%
,\cdots,\mathbf{Q}^{L-1|L},\mathbf{Q}^{L}\right)  $ is a generative model.

\subsection{2-Layer Soft Vector Quantiser (VQ) Network}

\label{Sect:2LayerFMC}The expression for $L\left(  \mathbf{P},\mathbf{Q}%
\right)  $ in equation \ref{Eq:ObjectiveMarkovSource} has a simple internal
structure which allows it to be systematically analysed. Thus apply equation
\ref{Eq:ObjectiveMarkovSource} to a 2-layer network where
\begin{align}
\mathbf{P}  & =\left(  \mathbf{P}^{0},\,\mathbf{P}^{1|0}\right) \nonumber\\
\mathbf{Q}  & =\left(  \mathbf{Q}^{0|1},\mathbf{Q}^{1}\right)
\end{align}
to obtain the objective function
\begin{equation}
L\left(  \mathbf{P},\mathbf{Q}\right)  =-\sum_{i_{0}=1}^{M_{0}}P_{i_{0}}%
^{0}\sum_{i_{1}=1}^{M_{1}}\,P_{i_{1},i_{0}}^{1|0}\,\log Q_{i_{0},i_{1}}%
^{0|1}+L\left(  \mathbf{P}^{1},\mathbf{Q}^{1}\right)
\label{Eq:ObjectiveMarkovSource2Layer}%
\end{equation}
Now change notation in order to make contact with previous results on vector
quantisers (VQ)
\begin{equation}%
\begin{tabular}
[c]{ll}%
$i_{0}\rightarrow\mathbf{x\;\;\;\;\;}\sum_{i_{0}=1}^{M_{0}}\rightarrow\int
d\mathbf{x}$ & input vector\\
$i_{1}\rightarrow y\mathbf{\;\;\;\;\;}\sum_{i_{1}=1}^{M_{1}}\rightarrow
\sum_{y=1}^{M}$ & output code index\\
$P_{i_{0}}^{0}\rightarrow\Pr\left(  \mathbf{x}\right)  $ & input PDF\\
$P_{i_{1},i_{0}}^{1|0}\rightarrow\Pr\left(  y|\mathbf{x}\right)  $ &
recognition model\\
$Q_{i_{0},i_{1}}^{0|1}\rightarrow V\frac{1}{\left(  \sqrt{2\pi}\sigma\right)
^{\dim\mathbf{x}}}\exp\left(  -\frac{\left\|  \mathbf{x}-\mathbf{x}^{\prime
}\left(  y\right)  \right\|  ^{2}}{2\sigma^{2}}\right)  $ & Gaussian
generative model\\
$Q_{i_{1}}^{1}\rightarrow Q\left(  y\right)  $ & output prior
\end{tabular}
\label{Eq:ObjectiveMarkovSource2LayerNotation}%
\end{equation}
where $\mathbf{x}$ is a continuous-valued input vector (e.g. the activity
pattern in layer 0), $\sigma$ is the (isotropic) variance of the Gaussian
generative model, $V$ is an infinitesimal volume element in input space which
may be used to convert the Gaussian probability density into a probability,
and $y$ is a discrete-valued output index (e.g. the location of the next
neuron to fire in layer 1). Note that the parameter $V$ must be introduced in
order to regularise the number of bits required to specify each source state.
In effect, $V$ specifies a resolution scale, such that details on smaller
scales are ignored.

The notation defined in equation \ref{Eq:ObjectiveMarkovSource2LayerNotation}
allows $L\left(  \mathbf{P},\mathbf{Q}\right)  $ to be written as
\begin{equation}
L\left(  \mathbf{P},\mathbf{Q}\right)  =\frac{D_{VQ}}{4\sigma^{2}}+L\left(
\mathbf{P}^{1},\mathbf{Q}^{1}\right)  -\log\left(  \frac{V}{\left(  \sqrt
{2\pi}\sigma\right)  ^{\dim\mathbf{x}}}\right)
\label{Eq:ObjectiveMarkovSource2Layer2}%
\end{equation}
where $D_{VQ}$ is defined as
\begin{equation}
D_{VQ}\equiv2\int d\mathbf{x}\Pr\left(  \mathbf{x}\right)  \sum_{y=1}^{M}%
\Pr\left(  y|\mathbf{x}\right)  \,\left\|  \mathbf{x}-\mathbf{x}^{\prime
}\left(  y\right)  \right\|  ^{2}\label{Eq:ObjectiveVQ}%
\end{equation}
The first term in equation \ref{Eq:ObjectiveMarkovSource2Layer2} is
proportional to the objective function $D_{VQ}$ for a soft vector quantiser
(VQ), where $\Pr\left(  y|\mathbf{x}\right)  $ is a soft encoder, and
$\mathbf{x}^{\prime}\left(  y\right)  $ is the corresponding reconstruction
vector attached to code index $y$, and$\,\left\|  \mathbf{x}-\mathbf{x}%
^{\prime}\left(  y\right)  \right\|  ^{2}$ is the $L^{2}$ norm of the
reconstruction error. A standard VQ \cite{Linde1980} (i.e. winner-take-all
encoder)\ has $\Pr\left(  y|\mathbf{x}\right)  =\delta_{y,y\left(
\mathbf{x}\right)  }$, which emerges as the optimal form when this VQ
objective function is minimised w.r.t. $\Pr\left(  y|\mathbf{x}\right)  $ (see
\cite{Luttrell1994a} for a detailed discussion of these issues). The second
term in equation \ref{Eq:ObjectiveMarkovSource2Layer2} (i.e. $L\left(
\mathbf{P}^{1},\mathbf{Q}^{1}\right)  $) is the cost of coding the output
layer, and the third term is constant.

The effect of the $L\left(  \mathbf{P}^{1},\mathbf{Q}^{1}\right)  $ term in
$L\left(  \mathbf{P},\mathbf{Q}\right)  $ (see equation
\ref{Eq:ObjectiveMarkovSource2Layer2})\ is to encourage $P_{i}^{1}%
\rightarrow\delta_{i,i_{0}}$ (only one state in layer 1 is used) and
$\mathbf{Q}^{1}\rightarrow\mathbf{P}^{1}$ (perfect model in layer 1). The
behaviour $P_{i}^{1}\rightarrow\delta_{i,i_{0}}$ is in conflict with the
requirements of the first term (i.e. the soft VQ) in $L\left(  \mathbf{P}%
,\mathbf{Q}\right)  $, which requires that more than one state in layer 1 is
used, in order to minimise the reconstruction distortion. There is a tradeoff
between increasing the number of active states in layer 1 in order to enable
the Gaussian generative model ($\mathbf{Q}^{0}$ is a Gaussian mixture
distribution) to make a good approximation to the external source
$\mathbf{P}^{0}$, and decreasing the number of active states in layer 1 in
order to make the average total number of bits $L\left(  \mathbf{P}%
^{1},\mathbf{Q}^{1}\right)  $ required to specify an output state as small as possible.

\subsection{Coupled Soft VQ Networks}

\label{Sect:CoupledFMC}The results of section \ref{Sect:2LayerFMC} will now be
generalised to an $\left(  L+1\right)  $-layer network. The objective function
for coding a Markov source (equation \ref{Eq:ObjectiveMarkovSource}) can be
written, using a notation which is analogous to that given in equation
\ref{Eq:ObjectiveMarkovSource2LayerNotation} as
\begin{equation}
L\left(  \mathbf{P},\mathbf{Q}\right)  =\sum_{l=0}^{L-1}\frac{D_{VQ}^{l}%
}{4\left(  \sigma_{l}\right)  ^{2}}+L\left(  \mathbf{P}^{L},\mathbf{Q}%
^{L}\right)  -\sum_{l=0}^{L-1}\log\left(  \frac{V_{l}}{\left(  \sqrt{2\pi
}\sigma_{l}\right)  ^{\dim\mathbf{x}_{l}}}\right)
\label{Eq:ObjectiveMarkovSourceSimple}%
\end{equation}
where $D_{VQ}^{l}$ is defined as ($D_{VQ}^{0}=D_{VQ}$ as defined in equation
\ref{Eq:ObjectiveVQ})
\begin{equation}
D_{VQ}^{l}\equiv2\int d\mathbf{x}_{l}\Pr\left(  \mathbf{x}_{l}\right)
\sum_{y_{l+1}=1}^{M_{l+1}}\Pr\left(  y_{l+1}|\mathbf{x}_{l}\right)  \,\left\|
\mathbf{x}_{l}-\mathbf{x}_{l}^{\prime}\left(  y_{l}\right)  \right\|  ^{2}%
\end{equation}
where $\mathbf{x}_{l}$ and $y_{l}$ are both used to denote the state of layer
$l$. The notation $\mathbf{x}_{l}$ is used to denote the input to the encoder
that connects layers $l$ and $l+1$, whereas the notation $y_{l}$ denotes the
output of the encoder that connects layers $l-1$ and $l$. This redundancy of
notation is not actually necessary, but is used here to preserve the
distinction between input vectors and output codes.

The first term in equation \ref{Eq:ObjectiveMarkovSourceSimple} is a weighted
sum (where each term is weighted by $\left(  \sigma_{l}\right)  ^{-2} $) of
objective functions for a set of soft VQs connecting each of the $L$
neighbouring pairs of layers in the network. This type of network structure
will be called a VQ-ladder. The second term in equation
\ref{Eq:ObjectiveMarkovSourceSimple} (i.e. $L\left(  \mathbf{P}^{L}%
,\mathbf{Q}^{L}\right)  $) is the cost of coding the output layer, and the
third term is constant.

If the cost $L\left(  \mathbf{P}^{L},\mathbf{Q}^{L}\right)  $ of coding the
output layer is ignored, then the multilayer Markov source coding objective
function $L\left(  \mathbf{P},\mathbf{Q}\right)  $ is minimised by minimising
the the objective function $\sum_{l=0}^{L-1}\frac{D_{VQ}^{l}}{\left(
\sigma_{l}\right)  ^{2}}$ for a VQ-ladder (see \cite{Luttrell1994a} discussion
of this point in the context of folded Markov chains (FMC)). As the number $L$
of network layers is increased, the effect of the $L\left(  \mathbf{P}%
^{L},\mathbf{Q}^{L}\right)  $ term has less and less effect on the overall
optimisation, because its effect is swamped by the VQ-ladder term.

\subsection{Topographic Mapping Network}

\label{Sect:Kohonen}The results obtained in section \ref{Sect:2LayerFMC} for a
soft VQ may be generalised to obtain a topographic mapping network whose
properties closely resemble those of a Kohonen network \cite{Kohonen1989}.
This derivation is based on the approach to topographic mappings that was
presented in \cite{Luttrell1989a}. Thus apply equation
\ref{Eq:ObjectiveMarkovSource} to a 3-layer network
\begin{align}
\mathbf{P}  & =\left(  \mathbf{P}^{0},\,\mathbf{P}^{1|0},\mathbf{P}%
^{2|1}\right) \nonumber\\
\mathbf{Q}  & =\left(  \mathbf{Q}^{0|1},\mathbf{Q}^{1|2},\mathbf{Q}%
^{2}\right)
\end{align}
where only layers $0$ and $2$ are included in the objective function, to
obtain
\begin{equation}
L\left(  \mathbf{P},\mathbf{Q}\right)  =-\sum_{i_{0}=1}^{M_{0}}P_{i_{0}}%
^{0}\sum_{i_{2}=1}^{M_{2}}\,P_{i_{2},i_{0}}^{2|0}\,\log Q_{i_{0},i_{2}}%
^{0|2}+L\left(  \mathbf{P}^{2},\mathbf{Q}^{2}\right)
\label{Eq:ObjectiveMarkovSource3Layer1Hidden}%
\end{equation}
which should be compared with equation \ref{Eq:ObjectiveMarkovSource2Layer}.
An analogous change of notation to that defined in equation
\ref{Eq:ObjectiveMarkovSource2LayerNotation} can be made
\begin{equation}%
\begin{tabular}
[c]{ll}%
$i_{0}\rightarrow\mathbf{x\;\;\;\;\;}\sum_{i_{0}=1}^{M_{0}}\rightarrow\int
d\mathbf{x}$ & input vector\\
$i_{1}\rightarrow y$ & hidden code index\\
$i_{2}\rightarrow z$ & output code index\\
$P_{i_{0}}^{0}\rightarrow\Pr\left(  \mathbf{x}\right)  $ & input PDF\\
$P_{i_{1},i_{0}}^{1|0}\rightarrow\Pr\left(  y|\mathbf{x}\right)  $ &
recognition model (first stage)\\
$P_{i_{2},i_{1}}^{2|1}\rightarrow\Pr\left(  z|y\right)  $ & recognition model
(second stage)\\
$Q_{i_{0},i_{2}}^{0|2}\rightarrow V\frac{1}{\left(  \sqrt{2\pi}\sigma\right)
^{\dim\mathbf{x}}}\exp\left(  -\frac{\left\|  \mathbf{x}-\mathbf{x}^{\prime
}\left(  z\right)  \right\|  ^{2}}{2\sigma^{2}}\right)  $ & Gaussian
generative model\\
$Q_{i_{2}}^{2}\rightarrow Q\left(  z\right)  $ & output prior
\end{tabular}
\end{equation}
to obtain
\begin{equation}
L\left(  \mathbf{P},\mathbf{Q}\right)  =\frac{D_{VQ}}{4\sigma^{2}}+L\left(
\mathbf{P}^{2},\mathbf{Q}^{2}\right)  -\log\left(  \frac{V}{\left(  \sqrt
{2\pi}\sigma\right)  ^{\dim\mathbf{x}}}\right)
\end{equation}
where $D_{VQ}$ is defined as
\begin{equation}
D_{VQ}\equiv2\int d\mathbf{x}\Pr\left(  \mathbf{x}\right)  \sum_{z=1}^{M_{2}%
}\Pr\left(  z|\mathbf{x}\right)  \,\left\|  \mathbf{x}-\mathbf{x}^{\prime
}\left(  z\right)  \right\|  ^{2}\label{Eq:ObjectiveVQ3Layer1Hidden}%
\end{equation}
which should be compared with the objective function in equation
\ref{Eq:ObjectiveVQ}.

This expression for $D_{VQ}$ explicitly involves the states of layers $0$ and
$2$ of a 3-layer network, and it will now be manipulated into a form that
explicitly involves the states of layers $0$ and $1$. In order to simplify
this calculation, $D_{VQ}$ will be replaced by the equivalent objective
function \cite{Luttrell1994a}
\begin{equation}
D_{VQ}\equiv\int d\mathbf{x}\Pr\left(  \mathbf{x}\right)  \sum_{z=1}^{M_{2}%
}\Pr\left(  z|\mathbf{x}\right)  \int d\mathbf{x}^{\prime}\Pr\left(
\mathbf{x}^{\prime}|z\right)  \,\left\|  \mathbf{x}-\mathbf{x}^{\prime
}\right\|  ^{2}%
\end{equation}
Now introduce dummy integrations over the state of layer $1$ to obtain
\begin{align}
D_{VQ}  & \equiv\int d\mathbf{x}\Pr\left(  \mathbf{x}\right)  \sum
_{y=1}^{M_{1}}\Pr\left(  y|\mathbf{x}\right)  \sum_{z=1}^{M_{2}}\Pr\left(
z|y\right)  \sum_{y^{\prime}=1}^{M_{1}}\Pr\left(  y^{\prime}|z\right)
\nonumber\\
& \times\int d\mathbf{x}^{\prime}\Pr\left(  \mathbf{x}^{\prime}|y^{\prime
}\right)  \,\left\|  \mathbf{x}-\mathbf{x}^{\prime}\right\|  ^{2}%
\end{align}
and rearrange to obtain
\begin{equation}
D_{VQ}\equiv\int d\mathbf{x}\Pr\left(  \mathbf{x}\right)  \sum_{y^{\prime}%
=1}^{M_{1}}\Pr\left(  y^{\prime}|\mathbf{x}\right)  \int d\mathbf{x}^{\prime
}\Pr\left(  \mathbf{x}^{\prime}|y^{\prime}\right)  \,\left\|  \mathbf{x}%
-\mathbf{x}^{\prime}\right\|  ^{2}%
\end{equation}
where
\begin{align}
\Pr\left(  y^{\prime}|y\right)   & =\sum_{z=1}^{M_{2}}\Pr\left(  y^{\prime
}|z\right)  \Pr\left(  z|y\right) \nonumber\\
\Pr\left(  y^{\prime}|\mathbf{x}\right)   & =\sum_{y=1}^{M_{1}}\Pr(y^{\prime
}|y)\Pr\left(  y|\mathbf{x}\right)
\end{align}
which may be replaced by the equivalent objective function
\begin{equation}
D_{VQ}\equiv2\int d\mathbf{x}\Pr\left(  \mathbf{x}\right)  \sum_{y^{\prime}%
=1}^{M_{1}}\Pr\left(  y^{\prime}|\mathbf{x}\right)  \,\left\|  \mathbf{x}%
-\mathbf{x}^{\prime}\left(  y^{\prime}\right)  \right\|  ^{2}%
\label{Eq:ObjectiveVQ3Layer1Hidden2}%
\end{equation}
which should be compared with the objective function in equation
\ref{Eq:ObjectiveVQ3Layer1Hidden}.

The overall effect of manipulating equation \ref{Eq:ObjectiveVQ3Layer1Hidden}
into the form given in equation \ref{Eq:ObjectiveVQ3Layer1Hidden2} is to
convert the objective function from one that explicitly involves the states of
layers $0$ and $2$, to one that explicitly involves the states of layers $0$
and $1$. This change is reflected in the replacement of $\Pr\left(
z|\mathbf{x}\right)  $ by $\Pr\left(  y^{\prime}|\mathbf{x}\right)  $. This
new form for the objective function (see equation
\ref{Eq:ObjectiveVQ3Layer1Hidden2}) is exactly the same as for a standard VQ
(see equation \ref{Eq:ObjectiveVQ}), except that the posterior probability
$\Pr\left(  y|\mathbf{x}\right)  $ is now processed through a transition
matrix $\Pr(y^{\prime}|y)$ to produce $\Pr\left(  y^{\prime}|\mathbf{x}%
\right)  $. Because $\Pr(y^{\prime}|y)=\sum_{z=1}^{M_{2}}\Pr\left(  y^{\prime
}|z\right)  \Pr\left(  z|y\right)  $, it takes account of the effect of the
state $z$ of layer $2$ on the training of layer $1$, which is a type of
self-supervision \cite{Luttrell1991b} in which higher layers of a network
coordinate the training of lower layers. However, viewed from the point of
view of layer $1$, the effect of the transition matrix $\Pr(y^{\prime}|y)$ is
to do damage to the posterior probability by redistributing probability
amongst the states of layer $1$. This process is thus called probability
leakage, and $\Pr(y^{\prime}|y)$ is called a probability leakage matrix.

The objective function in equation \ref{Eq:ObjectiveVQ3Layer1Hidden2} gives
rise to a neural network that closely resembles a Kohonen topographic mapping
neural network \cite{Kohonen1989}, where $\Pr(y^{\prime}|y)$ may be identified
as the topographic neighbourhood function, as was shown in
\cite{Luttrell1989a}. Note that in order for the topographic neighbourhood to
be localised (i.e. $\Pr(y^{\prime}|y)>0$ only for $y^{\prime}$ in some local
neighbourhood of $y$), the transition matrix $\Pr\left(  z|y\right)  $ that
generates the state of layer $2$ from the state of layer $1$ must generate
each $z$ state from $y$ states that are all close to each other. This
connection with Kohonen topographic mapping neural networks is only
approximate, because the training algorithm proposed by Kohonen does not
correspond to the minimisation of any objective function. A generalised
version of the Kohonen network which allows a factorial code to emerge may be
derived using the results in section \ref{Sect:PMD} \cite{Luttrell1993}.

\subsection{Additional Results}

\label{Sect:AdditionalResults}The objective function $\sum_{l=0}^{L-1}%
\frac{D_{VQ}^{l}}{4\left(  \sigma_{l}\right)  ^{2}}$ for a VQ-ladder couples
the optimisation of the individual 2-layer VQs together. Because the output of
the $l^{th}$ VQ is the input to the $\left(  l+1\right)  ^{th}$ VQ (for
$l=0,1,2\cdots,L-1\,$), the optimisation of the $k^{th}$ VQ has side effects
on the optimisation of the $l^{th}$ VQs (for $l=k+1,k+2,\cdots,L-1$). This
leads to the effect called self-supervision, in which top-down connections
from higher to lower network layers are automatically generated, to allow the
lower layers to process their input more effectively in the light of what the
higher layers discover in the data \cite{Luttrell1991b}. This is the
multilayer extension of the self-supervision effect that led to topographic
mappings in section \ref{Sect:Kohonen}.

The general expression for $L\left(  \mathbf{P},\mathbf{Q}\right)  $ in
equation \ref{Eq:ObjectiveMarkovSource} is the sum of two terms: the objective
function $\sum_{l=0}^{L-1}\sum_{i_{l}=1}^{M_{l}}P_{i_{l}}^{l}K_{i_{l}}\left(
\mathbf{P}^{l+1|l},\mathbf{Q}^{l|l+1}\right)  $ for a ladder (because
$\mathbf{Q}$ is not necessarily Gaussian, the ladder is not necessarily a
VQ-ladder), plus the cost $L\left(  \mathbf{P}^{L},\mathbf{Q}^{L}\right)  $ of
encoding layer $L$. The $L\left(  \mathbf{P}^{L},\mathbf{Q}^{L}\right)  $ term
has precisely the form that is commonly used in density modelling, so any
convenient density model could be used to parameterise $\mathbf{Q}^{L}$ in
layer $L$. A typical implementation of the type of network that minimises
$L\left(  \mathbf{P},\mathbf{Q}\right)  $ thus splits into two pieces
corresponding to the two different types of term in the objective function. In
the special case where $L=0$ (i.e. no ladder is used) this approach reduces to
standard input density modelling.

\section{Hierarachical Encoding using an Adaptive Cluster Expansion (ACE)}

\label{Sect:ACE}In this section the adaptive cluster expansion (ACE) network
is discussed \cite{Luttrell1989c}. ACE is a tree-structured network, whose
purpose is to decompose high-dimensional input vectors into a number of lower
dimensional pieces. In section \ref{Sect:ACETreeDensity} the case of a
deterministic source and a perfect model is considered, and in section
\ref{Sect:ACETreeDensityGaussian} the case of a Gaussian model is discussed.

\subsection{ACE:\ Tree-Structured Density Network}

\label{Sect:ACETreeDensity}Consider the objective function $L\left(
\mathbf{P},\mathbf{Q}\right)  $ for encoding an $L+1$ layer Markov source (see
equation \ref{Eq:ObjectiveMarkovSource}), and assume that the $Q_{i_{l}%
,i_{l+1}}^{l|l+1}$ part of the model is perfect so that $Q_{i_{l},i_{l+1}%
}^{l|l+1}=P_{i_{l},i_{l+1}}^{l|l+1}$ (for $l=0,1,\cdots,L-1$), and that the
$P_{i_{l+1},i_{l}}^{l+1|l}$ part of the source is deterministic so that
$P_{i_{l+1},i_{l}}^{l+1|l}=\delta_{i_{l+1},i_{l+1}\left(  i_{l}\right)  }$
(for $l=0,1,\cdots,L-1$), in which case $L\left(  \mathbf{P},\mathbf{Q}%
\right)  $ simplifies as follows (see appendix \ref{Appendix:ACEnonTree})
\begin{equation}
L\left(  \mathbf{P},\mathbf{Q}\right)  =H\left(  \mathbf{P}^{0}\right)
-H\left(  \mathbf{P}^{L}\right)  +L\left(  \mathbf{P}^{L},\mathbf{Q}%
^{L}\right) \label{Eq:ObjectiveACE1}%
\end{equation}
where $H\left(  \mathbf{P}^{0}\right)  -H\left(  \mathbf{P}^{L}\right)  $ is
the number of bits per symbol required to convert a $\mathbf{P}^{L}$-message
into a $\mathbf{P}^{0}$-message, assuming that the $P_{i_{l+1},i_{l}}^{l+1|l}
$ part of the source is deterministic, and that the model is perfect. This
result is not very interesting in itself.

However, if the $P_{i_{l+1},i_{l}}^{l+1|l}$ part of the source is not only
deterministic, but is also tree-structured, and the model is similarly
tree-structured, then the notation must be modified thus
\begin{align}
i_{l}  & \rightarrow\mathbf{i}_{l}=\left(  \mathbf{i}_{l}^{1},\mathbf{i}%
_{l}^{2},\cdots\right) \nonumber\\
i_{l+1}  & \rightarrow\mathbf{i}_{l+1}=\left(  i_{l+1}^{1},i_{l+1}^{2}%
,\cdots\right) \nonumber\\
P_{i_{l+1},i_{l}}^{l+1|l}  & \rightarrow P_{\mathbf{i}_{l+1},\mathbf{i}_{l}%
}^{l+1|l}=P_{i_{l+1}^{1},\mathbf{i}_{l}^{1}}^{l+1|l}P_{i_{l+1}^{2}%
,\mathbf{i}_{l}^{2}}^{l+1|l}\cdots=\delta_{i_{l+1}^{1},i_{l+1}^{1}\left(
\mathbf{i}_{l}^{1}\right)  }\delta_{i_{l+1}^{2},i_{l+1}^{2}\left(
\mathbf{i}_{l}^{2}\right)  }\cdots\nonumber\\
Q_{i_{l},i_{l+1}}^{l|l+1}  & \rightarrow Q_{\mathbf{i}_{l},\mathbf{i}_{l+1}%
}^{l|l+1}=P_{\mathbf{i}_{l}^{1},i_{l+1}^{1}}^{l|l+1}P_{\mathbf{i}_{l}%
^{2},i_{l+1}^{2}}^{l|l+1}\cdots\label{Eq:NotationACE}%
\end{align}
where the state $i_{l}$ of layer $l$ of the tree-structured Markov source is
more naturally written as a vector state $\mathbf{i}_{l}$ that specifies the
joint state of each branch of layer $l$ of the tree (the $i_{l}$ style of
notation is more suitable for a non-tree-structured Markov source).
Furthermore, the components of the vector $\mathbf{i}_{l}$ are partitioned as
$\left(  \mathbf{i}_{l}^{1},\mathbf{i}_{l}^{2},\cdots\right)  $, where each
$\mathbf{i}_{l}^{c}$ is the joint state of a subset $c$ of nodes in layer $l$,
where all the nodes in each subset are all siblings as seen from the point of
view of layer $l+1$. Such a set of siblings is called a cluster. The
components of the vector $\mathbf{i}_{l+1}$ are partitioned as $\left(
i_{l+1}^{1},i_{l+1}^{2},\cdots\right)  $, where $i_{l+1}^{c}$ is the state of
the parent (in layer $l+1$) of the siblings in cluster $c$ in layer $l$.

This notation may be used to rearrange $L\left(  \mathbf{P},\mathbf{Q}\right)
$ as follows (see appendix \ref{Appendix:ACEtree})
\begin{equation}
L\left(  \mathbf{P},\mathbf{Q}\right)  =\sum_{l=0}^{L-1}\sum_{\text{cluster
}c}H\left(  \mathbf{P}_{c}^{l}\right)  -\sum_{l=1}^{L}\sum_{\text{component
}c}H\left(  \mathbf{P}_{c}^{l}\right)  +L\left(  \mathbf{P}^{L},\mathbf{Q}%
^{L}\right) \label{Eq:ObjectiveACE2}%
\end{equation}
This expression for $L\left(  \mathbf{P},\mathbf{Q}\right)  $ can be rewritten
in terms of the mutual information $I\left(  \mathbf{P}_{c}^{l}\right)  $
between the components of cluster $\mathbf{i}_{l+1}^{c}$ as (see appendix
\ref{Appendix:ACEtree})
\begin{align}
L\left(  \mathbf{P},\mathbf{Q}\right)   & =-\sum_{l=1}^{L}\sum_{\text{cluster
}c}I\left(  \mathbf{P}_{c}^{l}\right)  +\sum_{\text{cluster }c}H\left(
\mathbf{P}_{c}^{0}\right)  -\sum_{\text{cluster }c}H\left(  \mathbf{P}_{c}%
^{L}\right) \nonumber\\
& +L\left(  \mathbf{P}^{L},\mathbf{Q}^{L}\right)
\end{align}
Now assume that the model is perfect in the output layer, so that
$\mathbf{Q}^{L}$ is given by $Q_{\mathbf{i}_{L}}^{L}=P_{\mathbf{i}_{l}^{1}%
}^{L}P_{\mathbf{i}_{l}^{2}}^{L}\cdots$. This allows $L\left(  \mathbf{P}%
^{L},\mathbf{Q}^{L}\right)  $ to be simplified as $L\left(  \mathbf{P}%
^{L},\mathbf{Q}^{L}\right)  =\sum_{\text{cluster }c}H\left(  \mathbf{P}%
_{c}^{L}\right)  $, so that $L\left(  \mathbf{P},\mathbf{Q}\right)  $ may
finally be expressed as
\begin{equation}
L\left(  \mathbf{P},\mathbf{Q}\right)  =-\sum_{l=1}^{L}\sum_{\text{cluster }%
c}I\left(  \mathbf{P}_{c}^{l}\right)  +\sum_{\text{cluster }c}H\left(
\mathbf{P}_{c}^{0}\right)
\end{equation}
The $-\sum_{l=1}^{L}\sum_{\text{cluster }c}I\left(  \mathbf{P}_{c}^{l}\right)
$ term is (minus) the sum of the mutual informations within all of the
clusters in the $L+1$ layer network, and the $\sum_{\text{cluster }c}H\left(
\mathbf{P}_{c}^{0}\right)  $ term is constant for a given external source
$\mathbf{P}^{0}$. This means that minimising $L\left(  \mathbf{P}%
,\mathbf{Q}\right)  $ is equivalent to maximising $\sum_{l=1}^{L}%
\sum_{\text{cluster }c}I\left(  \mathbf{P}_{c}^{l}\right)  $. This is the
maximum mutual information result for ACE networks \cite{Luttrell1991a}, which
includes the mutual information maximisation principle in \cite{Becker1992} as
a special case.

Note that if the source is deterministic and the model is perfect (as they are
here), then $L\left(  \mathbf{P}^{0},\mathbf{Q}^{0}\right)  =L\left(
\mathbf{P},\mathbf{Q}\right)  $, which implies that input density optimisation
is equivalent to joint density optimisation. This equivalence was used in
\cite{Luttrell1991a}, where the sum-of-mutual-informations objective function
was derived by minimising $L\left(  \mathbf{P}^{0},\mathbf{Q}^{0}\right)  $.

\subsection{ACE:\ Hierarchical Vector Quantiser}

\label{Sect:ACETreeDensityGaussian}If the above ACE\ network is modified
slightly, so that the model $\mathbf{Q}$ has exactly the same structure as
before, but is Gaussian rather than perfect, then $Q_{i_{l},i_{l+1}}^{l|l+1}$
becomes
\begin{equation}
Q_{i_{l},i_{l+1}}^{l|l+1}\rightarrow Q_{\mathbf{i}_{l},\mathbf{i}_{l+1}%
}^{l|l+1}=Q_{\mathbf{i}_{l}^{1},i_{l+1}^{1}}^{l|l+1}Q_{\mathbf{i}_{l}%
^{2},i_{l+1}^{2}}^{l|l+1}\cdots
\end{equation}
where the individual $Q_{\mathbf{i}_{l}^{c},i_{l+1}^{c}}^{l|l+1}$ are
Gaussian. The expression for $L\left(  \mathbf{P},\mathbf{Q}\right)  $ may
then be written down by analogy with equation
\ref{Eq:ObjectiveMarkovSourceSimple}
\begin{align}
L\left(  \mathbf{P},\mathbf{Q}\right)   & =\sum_{l=0}^{L-1}\sum_{\text{cluster
}c}\frac{D_{VQ}^{l,c}}{4\left(  \sigma_{l,c}\right)  ^{2}}+L\left(
\mathbf{P}^{L},\mathbf{Q}^{L}\right) \nonumber\\
& -\sum_{l=0}^{L-1}\sum_{\text{cluster }c}\log\left(  \frac{V_{l,c}}{\left(
\sqrt{2\pi}\sigma_{l,c}\right)  ^{\dim\mathbf{i}_{l}^{c}}}\right)
\end{align}
Thus the ACE\ network, with a Gaussian model $\mathbf{Q}$, is a hierarchical
VQ-ladder (or VQ-tree), in which each layer encodes the clusters in the
previous layer \cite{Luttrell1989a}.

\section{Factorial Encoding using a Partitioned Mixture Distribution (PMD)}

\label{Sect:PMD}In this section a useful parameterisation of the conditional
probability $\mathbf{P}^{l+1|l}$ for building the Markov source is introduced
in order to encourage $\mathbf{P}^{l+1|l}$ to form factorial codes of the
state of layer $l$. It turns out that there is a simple way of allowing such
codes to develop, which is called the partitioned mixture distribution (PMD)
\cite{Luttrell1993}. A PMD achieves this by encoding its input simultaneously
with a number of different recognition models, each of which potentially can
encode a different part of the input.

\bigskip In section \ref{Sect:MultipleRecognition} two ways in which multiple
recognition models can be used for factorial encoding are discussed, and a
hybrid approach (which is a PMD) is discussed in section
\ref{Sect:MultipleRecognitionAverage}.

\subsection{Multiple Recognition Models}

\label{Sect:MultipleRecognition}In the expression for the $L\left(
\mathbf{P},\mathbf{Q}\right)  $ (see equation \ref{Eq:ObjectiveMarkovSource})
the generative models $\mathbf{Q}^{l|l+1}$ may be parameterised as Gaussian
probability densities, whereas the recognition models $\mathbf{P}^{l+1|l}$ may
be parameterised in a more general way as
\begin{equation}
P_{i_{l+1},i_{l}}^{l+1|l}=\frac{P_{i_{l},i_{l+1}}^{l|l+1}P_{i_{l+1}}^{l+1}%
}{\sum_{i_{l+1}^{\prime}=1}^{M_{l+1}}P_{i_{l},i_{l+1}^{\prime}}^{l|l+1}%
P_{i_{l+1}^{\prime}}^{l+1}}\label{Eq:PosteriorProbability}%
\end{equation}
which guarantees the normalisation condition $\sum_{i_{l+1}=1}^{M_{l+1}%
}P_{i_{l+1},i_{l}}^{l+1|l}=1$. A limitation of this type of recognition model
is that it allows only a single explanation $i_{l+1}$ of the data $i_{l}$ (in
the case of a hard $P_{i_{l+1},i_{l}}^{l+1|l}$), or a probability distribution
over single explanations (in the case of a soft $P_{i_{l+1},i_{l}}^{l+1|l}$),
so it cannot lead to a factorial encoding of the data.

The simplest way of allowing a factorial encoding to develop is to make
simultaneous use more than one recognition model. Each recognition model uses
its own $\mathbf{P}^{l+1}$ vector and $\mathbf{P}^{l|l+1}$ matrix to compute a
posterior probability of the type shown in equation
\ref{Eq:PosteriorProbability}, so that if each recognition model is sensitised
to a different part of the input, then a factorial code can develop. This
approach can be formalised by making the replacement $i_{l+1}\rightarrow
\mathbf{i}_{l+1}$ in equation \ref{Eq:PosteriorProbability} (i.e. replace the
scalar code index by a vector code index, where the number of vector
components is equal to the number of recognition models). If the components of
$\mathbf{i}_{l+1}$ are determined independently of each other, then their
joint posterior probability $P_{\mathbf{i}_{l+1},i_{l}}^{l+1|l}$ is a product
of independent posterior probabilities, where each posterior probability
corresponds to one of the recognition models, and thus has its own
$\mathbf{P}^{l+1}$ vector and $\mathbf{P}^{l|l+1}$ matrix.

If this type of posterior probability, which is a product of $n$ independent
factors if there are $n$ independent recognition models, is then inserted into
equation \ref{Eq:ObjectiveVQ}\ it yields (see appendix \ref{Appendix:PMD})
\begin{align}
D_{VQ}  & \leq2\int d\mathbf{x}\Pr\left(  \mathbf{x}\right)  \sum_{y_{1}%
=1}^{M_{1}}\sum_{y_{2}=1}^{M_{2}}\cdots\sum_{y_{n}=1}^{M_{n}}\Pr\left(
y_{1}|\mathbf{x},1\right)  \Pr\left(  y_{2}|\mathbf{x},2\right)
\cdots\nonumber\\
& \cdots\Pr\left(  y_{n}|\mathbf{x},n\right)  \left\|  \mathbf{x}-\frac{1}%
{n}\sum_{k=1}^{n}\mathbf{x}_{k}^{\prime}\left(  y_{k}\right)  \right\|
^{2}\label{Eq:MultipleRecognition}%
\end{align}
If a single recognition model is independently used $n$ times, rather than $n
$ independent recognition models each independently being used once, then the
above result becomes
\begin{align}
D_{VQ}  & \leq\frac{2}{n}\int d\mathbf{x}\Pr\left(  \mathbf{x}\right)
\sum_{y=1}^{M}\Pr\left(  y|\mathbf{x}\right)  \,\left\|  \mathbf{x}%
-\mathbf{x}^{\prime}\left(  y\right)  \right\|  ^{2}\nonumber\\
& +\frac{2\left(  n-1\right)  }{n}\int d\mathbf{x}\Pr\left(  \mathbf{x}%
\right)  \left\|  \mathbf{x}-\sum_{y=1}^{M}\Pr\left(  y|\mathbf{x}\right)
\mathbf{x}^{\prime}\left(  y\right)  \right\|  ^{2}%
\label{Eq:SingleRecognition}%
\end{align}
In the case $n=1$ this correctly reduces to equation \ref{Eq:ObjectiveVQ} (the
inequality reduces to an equality in this case). When $n>1\,$the second term
offers the possibility of factorial encoding, because it contains a weighted
linear combination $\sum_{y=1}^{M}\Pr\left(  y|\mathbf{x}\right)
\mathbf{x}^{\prime}\left(  y\right)  $ of vectors.

\subsection{Average Over Recognition Models}

\label{Sect:MultipleRecognitionAverage}Now combine the above two approaches to
factorial encoding, so that a single recognition model is used (as in equation
\ref{Eq:SingleRecognition}), which is parameterised in such a way that it can
emulate multiple recognition models (as in equation
\ref{Eq:MultipleRecognition}). The simplest possibility is to firstly make the
replacement $P_{i_{l+1}}^{l+1}\rightarrow A_{k,i_{l+1}}^{l+1}P_{i_{l+1}}%
^{l+1}$ (where $A_{k,i_{l+1}}^{l+1}\geq0$)\ in equation
\ref{Eq:PosteriorProbability}, where $k$ is a recognition model index which
ranges over $k=1,2,\cdots,K$ (note that $K$ is not constrained to be the same
as $n$), and then secondly to average over $k$, to produce
\begin{equation}
P_{i_{l+1},i_{l}}^{l+1|l}\rightarrow\frac{1}{K}\sum_{k=1}^{K}\frac
{P_{i_{l},i_{l+1}}^{l|l+1}A_{k,i_{l+1}}^{l+1}P_{i_{l+1}}^{l+1}}{\sum
_{i_{l+1}^{\prime}=1}^{M_{l+1}}P_{i_{l},i_{l+1}^{\prime}}^{l|l+1}%
A_{k,i_{l+1}^{\prime}}^{l+1}P_{i_{l+1}^{\prime}}^{l+1}}%
\label{Eq:PosteriorProbabilityPMD}%
\end{equation}
In effect, $K$ recognition models are embedded between layer $l$ and layer
$l+1$, and the $\mathbf{A}^{l+1}$ matrix specifies which indices $i_{l+1}$ in
layer $l+1$ are associated with recognition model $k$.

The result in equation \ref{Eq:PosteriorProbabilityPMD} is not the same as the
result that would have been obtained using a Bayesian analysis, in which the
posterior probabilities generated by different models are combined to yield a
single posterior probability. In appendix \ref{Appendix:PMD} there is a
discussion of the relationship between the above proposed PMD recognition
model and a full Bayesian average over alternative recognition models.

A partitioned mixture distribution (PMD) is precisely this type of multiple
embedded recognition model. In the simplest type of PMD the $\mathbf{A}^{l+1}
$ matrix is chosen to contain only 0's and 1's, which are arranged so that the
$K$ recognition models partition layer $l+1$ into $K$ overlapping patches
\cite{Luttrell1993}. A wide range of types of PMD can be constructed by
choosing $\mathbf{A}^{l+1}$ appropriately.

In section \ref{Sect:Kohonen} it was shown how a Kohonen topographic mapping
emerged when a 3-layer Markov source network was optimised. If the PMD
posterior probability (see equation \ref{Eq:PosteriorProbabilityPMD}) had been
used in section \ref{Sect:Kohonen}, then a more general form of topographic
mapping (i.e. a factorial topographic mapping) would have emerged (this is
briefly discussed in \cite{Luttrell1993}).

\section{Conclusions}

The objective function for optimising the density model of a Markov source may
be applied to the problem of optimising the joint density of all the layers of
a neural network. This is possible because the joint state of all of the
network layers may be viewed as a Markov chain of states (each layer is
connected only to adjacent layers). This representation makes contact with the
results that were reported in \cite{Luttrell1994a}, and allows many results to
be unified into a single approach (i.e. a single objective function).

The most significant aspect of this unification is the fact that all layers of
a neural network are treated on an equal footing, unlike in the conventional
approach to density modelling where the input layer is accorded a special
status. For instance, this leads to a modular approach to building neural
networks, where all of the modules have the same structure.

\section{Acknowledgements}

I thank Chris Webber for many useful conversations that we had during the
course of this research. I also thank Peter Dayan and Geoffrey Hinton for
conversations that we had about the relationship between Markov source
modelling and Helmholtz machines during the 1997 ``Neural Networks and Machine
Learning'' programme at the Newton Institute in Cambridge.

\appendix

\section{ACE}

\label{Appendix:ACE}In this appendix some of the more technical details
relevant to section \ref{Sect:ACE} are given.

\subsection{Perfect Model, Deterministic Source}

\label{Appendix:ACEnonTree}The derivation of the result in equation
\ref{Eq:ObjectiveACE1} for a perfect model (i.e. $\mathbf{Q=P}$) and a
deterministic source (i.e. $P_{i_{l+1},i_{l}}^{l+1|l}=\delta_{i_{l+1}%
,i_{l+1}\left(  i_{l}\right)  }$ using scalar notation $i_{l}$ rather than
vector notation $\mathbf{i}_{l}$ for the state of layer $l$, because here the
Markov source is not assumed to be tree-structured) is as follows. The basic
definition of $L\left(  \mathbf{P},\mathbf{Q}\right)  $ in equation
\ref{Eq:ObjectiveMarkovSource} may be written as
\begin{equation}
L\left(  \mathbf{P},\mathbf{Q}\right)  -L\left(  \mathbf{P}^{L},\mathbf{Q}%
^{L}\right)  =-\sum_{l=0}^{L-1}\sum_{i_{l}=1}^{M_{l}}P_{i_{l}}^{l}%
\sum_{i_{l+1}=1}^{M_{l+1}}\,P_{i_{l+1},i_{l}}^{l+1|l}\,\log P_{i_{l},i_{l+1}%
}^{l|l+1}%
\end{equation}
This may be simplified by noting that $P_{i_{l+1},i_{l}}^{l+1|l}%
=\delta_{i_{l+1},i_{l+1}\left(  i_{l}\right)  }$, and that Bayes' theorem
gives $P_{i_{l},i_{l+1}}^{l|l+1}=\frac{P_{i_{l+1},i_{l}}^{l+1|l}P_{i_{l}}^{l}%
}{P_{i_{l+1}}^{l+1}}$, which yields
\begin{equation}
L\left(  \mathbf{P},\mathbf{Q}\right)  -L\left(  \mathbf{P}^{L},\mathbf{Q}%
^{L}\right)  =-\sum_{l=0}^{L-1}\sum_{i_{l}=1}^{M_{l}}P_{i_{l}}^{l}%
\sum_{i_{l+1}=1}^{M_{l+1}}\,\delta_{i_{l+1},i_{l+1}\left(  i_{l}\right)
}\,\log\frac{\delta_{i_{l+1},i_{l+1}\left(  i_{l}\right)  }P_{i_{l}}^{l}%
}{P_{i_{l+1}}^{l+1}}%
\end{equation}
Now use that $\sum_{i_{l+1}=1}^{M_{l+1}}\,\delta_{i_{l+1},i_{l+1}\left(
i_{l}\right)  }=1$, $\sum_{i_{l+1}=1}^{M_{l+1}}\,\delta_{i_{l+1}%
,i_{l+1}\left(  i_{l}\right)  }\,\log\delta_{i_{l+1},i_{l+1}\left(
i_{l}\right)  }=0$ and $\sum_{i_{l}=1}^{M_{l}}P_{i_{l}}^{l}\delta
_{i_{l+1},i_{l+1}\left(  i_{l}\right)  }=P_{i_{l+1}}^{l+1}$ to reduce this to
\begin{equation}
L\left(  \mathbf{P},\mathbf{Q}\right)  -L\left(  \mathbf{P}^{L},\mathbf{Q}%
^{L}\right)  =-\sum_{l=0}^{L-1}\sum_{i_{l}=1}^{M_{l}}P_{i_{l}}^{l}\,\log
P_{i_{l}}^{l}+\sum_{l=0}^{L-1}\sum_{i_{l+1}=1}^{M_{l+1}}\,P_{i_{l+1}}%
^{l+1}\log P_{i_{l+1}}^{l+1}%
\end{equation}
The terms in these two series mostly cancel each other to yield
\begin{equation}
L\left(  \mathbf{P},\mathbf{Q}\right)  -L\left(  \mathbf{P}^{L},\mathbf{Q}%
^{L}\right)  =-\sum_{i_{0}=1}^{M_{0}}P_{i_{0}}^{0}\,\log P_{i_{0}}^{0}%
+\sum_{i_{L}=1}^{M_{L}}\,P_{i_{L}}^{L}\log P_{i_{_{L}}}^{L}%
\end{equation}
and using the definition of entropy (see equation \ref{Eq:Entropy}) this may
finally be written as
\begin{equation}
L\left(  \mathbf{P},\mathbf{Q}\right)  -L\left(  \mathbf{P}^{L},\mathbf{Q}%
^{L}\right)  =H\left(  \mathbf{P}^{0}\right)  -H\left(  \mathbf{P}^{L}\right)
\end{equation}

\subsection{Perfect Model, Deterministic Source: Tree-Structured Case}

\label{Appendix:ACEtree}The derivation of the result in equation
\ref{Eq:ObjectiveACE2} for a perfect tree-structured model and a deterministic
tree-structured source is may be obtained by altering the notation in appendix
\ref{Appendix:ACEnonTree} to reflect the fact that both the Markov source and
model are now tree-structured. Thus use the notation defined in equation
\ref{Eq:NotationACE} to write $L\left(  \mathbf{P},\mathbf{Q}\right)  $ (see
equation \ref{Eq:ObjectiveMarkovSource}) as%

\begin{equation}
L\left(  \mathbf{P},\mathbf{Q}\right)  -L\left(  \mathbf{P}^{L},\mathbf{Q}%
^{L}\right)  =-\sum_{l=0}^{L-1}\sum_{\mathbf{i}_{l}}P_{\mathbf{i}_{l}}^{l}%
\sum_{\mathbf{i}_{l+1}}\,P_{\mathbf{i}_{l+1},\mathbf{i}_{l}}^{l+1|l}%
\,\sum_{\text{cluster }c}\log P_{\mathbf{i}_{l}^{c},i_{l+1}^{c}}^{l|l+1}%
\end{equation}
Now use Bayes' theorem in the form $P_{\mathbf{i}_{l}^{c},i_{l+1}^{c}}%
^{l|l+1}=\frac{P_{i_{l+1}^{c},\mathbf{i}_{l}^{c}}^{l+1|l}P_{\mathbf{i}_{l}%
^{c}}^{l}}{P_{i_{l+1}^{c}}^{l+1}}$ to write this as
\begin{equation}
L\left(  \mathbf{P},\mathbf{Q}\right)  -L\left(  \mathbf{P}^{L},\mathbf{Q}%
^{L}\right)  =-\sum_{l=0}^{L-1}\sum_{\mathbf{i}_{l}}P_{\mathbf{i}_{l}}^{l}%
\sum_{\mathbf{i}_{l+1}}\,P_{\mathbf{i}_{l+1},\mathbf{i}_{l}}^{l+1|l}%
\,\sum_{\text{cluster }c}\log\left(  \frac{P_{i_{l+1}^{c},\mathbf{i}_{l}^{c}%
}^{l+1|l}P_{\mathbf{i}_{l}^{c}}^{l}}{P_{i_{l+1}^{c}}^{l+1}}\right)
\end{equation}
This may be simplified by using that $\sum_{\mathbf{i}_{l+1}}\,P_{\mathbf{i}%
_{l+1},\mathbf{i}_{l}}^{l+1|l}=1$ (for the $\log P_{\mathbf{i}_{l}^{c}}^{l}$
term), $\sum_{\mathbf{i}_{l}}P_{\mathbf{i}_{l}}^{l}\sum_{\mathbf{i}_{l+1}%
}\,P_{\mathbf{i}_{l+1},\mathbf{i}_{l}}^{l+1|l}\left(  \cdots\right)
=\sum_{\mathbf{i}_{l+1}}P_{\mathbf{i}_{l+1}}^{l+1}\left(  \cdots\right)  $
(for the $\log P_{i_{l+1}^{c}}^{l+1}$ term), and $\sum_{i_{l+1}=1}^{M_{l+1}%
}\,\delta_{i_{l+1}^{c},i_{l+1}^{c}\left(  \mathbf{i}_{l}^{c}\right)  }%
\,\log\delta_{i_{l+1}^{c},i_{l+1}^{c}\left(  \mathbf{i}_{l}^{c}\right)  }=0$
(for the $\log P_{i_{l+1}^{c},\mathbf{i}_{l}^{c}}^{l+1|l}$ term), to yield
\begin{align}
L\left(  \mathbf{P},\mathbf{Q}\right)  -L\left(  \mathbf{P}^{L},\mathbf{Q}%
^{L}\right)   & =-\sum_{l=0}^{L-1}\sum_{\mathbf{i}_{l}}P_{\mathbf{i}_{l}}%
^{l}\sum_{\text{cluster }c}\log P_{\mathbf{i}_{l}^{c}}^{l}\nonumber\\
& +\sum_{l=0}^{L-1}\sum_{\mathbf{i}_{l+1}}\,P_{\mathbf{i}_{l+1}}^{l+1}%
\,\sum_{\text{cluster }c}\log P_{i_{l+1}^{c}}^{l+1}%
\end{align}%
\sloppypar
The first term may be simplified by interchanging the order of summation
$\sum_{\mathbf{i}_{l}}\sum_{c}\left(  \cdots\right)  =\sum_{c}\sum
_{\mathbf{i}_{l}}\left(  \cdots\right)  $, and then marginalising the
probabilities using that $\sum_{\text{cluster }c}\sum_{\mathbf{i}_{l}%
}P_{\mathbf{i}_{l}}^{l}\log P_{\mathbf{i}_{l}^{c}}^{l}=\sum_{\text{cluster }%
c}\sum_{\mathbf{i}_{l}^{c}}P_{\mathbf{i}_{l}^{c}}^{l}\log P_{\mathbf{i}%
_{l}^{c}}^{l}$. The second term may be simplified by interchanging the order
of summation $\sum_{\mathbf{i}_{l+1}}\sum_{\text{cluster }c}\left(
\cdots\right)  =\sum_{\text{cluster }c}\sum_{\mathbf{i}_{l+1}}\left(
\cdots\right)  $, then marginalising the probabilities using that
$\sum_{\text{cluster }c}\sum_{\mathbf{i}_{l+1}}P_{\mathbf{i}_{l+1}}^{l+1}\log
P_{i_{l+1}^{c}}^{l+1}=\sum_{\text{cluster }c}\sum_{i_{l+1}^{c}}P_{i_{l+1}^{c}%
}^{l+1}\log P_{i_{l+1}^{c}}^{l+1}$, and then using that component $c$ in layer
$l+1$ is the parent of cluster $c$ in layer $l$, to obtain
\begin{align}
L\left(  \mathbf{P},\mathbf{Q}\right)  -L\left(  \mathbf{P}^{L},\mathbf{Q}%
^{L}\right)   & =-\sum_{l=0}^{L-1}\sum_{\mathbf{i}_{l}^{c}}P_{\mathbf{i}%
_{l}^{c}}^{l}\sum_{\text{cluster }c}\log P_{\mathbf{i}_{l}^{c}}^{l}\nonumber\\
& +\sum_{l=1}^{L}\sum_{i_{l}^{c}}P_{i_{l}^{c}}^{l}\,\sum_{\text{component }%
c}\log P_{i_{l}^{c}}^{l}%
\end{align}
and using the definition of entropy (see equation \ref{Eq:Entropy}) this may
finally be written as
\begin{equation}
L\left(  \mathbf{P},\mathbf{Q}\right)  -L\left(  \mathbf{P}^{L},\mathbf{Q}%
^{L}\right)  =\sum_{l=0}^{L-1}\sum_{\text{cluster }c}H\left(  \mathbf{P}%
_{c}^{l}\right)  -\sum_{l=1}^{L}\sum_{\text{component }c}H\left(  P_{c}%
^{l}\right)
\end{equation}
where $H\left(  \mathbf{P}_{c}^{l}\right)  $ is the entropy of cluster $c$ and
$H\left(  P_{c}^{l}\right)  $ is the entropy of component $c$ (both in layer
$l $). The mutual information $I\left(  \mathbf{P}_{c}^{l}\right)  $ between
the components $c^{\prime}$ of cluster $c$ is defined as
\begin{equation}
I\left(  \mathbf{P}_{c}^{l}\right)  \equiv\sum_{\substack{\text{component
}c^{\prime} \\\text{in cluster }c }}H\left(  P_{c^{\prime}}^{l}\right)
-H\left(  \mathbf{P}_{c}^{l}\right)
\end{equation}
and using that $\sum_{\text{cluster }c}\sum_{\substack{\text{component
}c^{\prime} \\\text{in cluster }c }}\left(  \cdots\right)  =\sum
_{\text{component }c^{\prime}}$ this yields%

\begin{equation}
\sum_{\text{cluster }c}I\left(  \mathbf{P}_{c}^{l}\right)  \equiv
\sum_{\text{component }c}H\left(  P_{c}^{l}\right)  -\sum_{\text{cluster }%
c}H\left(  \mathbf{P}_{c}^{l}\right)
\end{equation}
which allows $L\left(  \mathbf{P},\mathbf{Q}\right)  -L\left(  \mathbf{P}%
^{L},\mathbf{Q}^{L}\right)  $ to be simplified to
\begin{align}
L\left(  \mathbf{P},\mathbf{Q}\right)  -L\left(  \mathbf{P}^{L},\mathbf{Q}%
^{L}\right)   & =-\sum_{l=1}^{L}\sum_{\text{cluster }c}I\left(  \mathbf{P}%
_{c}^{l}\right) \nonumber\\
& +\sum_{\text{cluster }c}H\left(  \mathbf{P}_{c}^{0}\right)  -\sum
_{\text{cluster }c}H\left(  \mathbf{P}_{c}^{L}\right)
\end{align}

\section{PMD}

\label{Appendix:PMD}In this appendix some of the more technical details
relevant to section \ref{Sect:PMD} are given.

\subsection{PMD Recognition Model}

\label{Appendix:PMDRecognition}If the type of posterior probability introduced
in section \ref{Sect:MultipleRecognition}, which is a product of $n$
independent factors if there are $n$ independent recognition models, is then
inserted into equation \ref{Eq:ObjectiveVQ}\ it yields a $D_{VQ}$ of the form
\begin{align}
D_{VQ}  & =2\int d\mathbf{x}\Pr\left(  \mathbf{x}\right)  \sum_{y_{1}%
=1}^{M_{1}}\sum_{y_{2}=1}^{M_{2}}\cdots\sum_{y_{n}=1}^{M_{n}}\Pr\left(
y_{1}|\mathbf{x},1\right)  \Pr\left(  y_{2}|\mathbf{x},2\right)
\cdots\,\nonumber\\
& \cdots\Pr\left(  y_{n}|\mathbf{x},n\right)  \left\|  \mathbf{x}%
-\mathbf{x}^{\prime}\left(  y_{1},y_{2},\cdots,y_{n}\right)  \right\|  ^{2}%
\end{align}
where $\Pr\left(  y_{k}|\mathbf{x},k\right)  $ denotes the posterior
probability that (given input $\mathbf{x}$) code index $y_{k}$ occurs in
recognition model $k$. If $\mathbf{x}^{\prime}\left(  y_{1},y_{2},\cdots
,y_{n}\right)  $ is optimised (i.e. takes the value that minimises $D_{VQ}$)
then it becomes
\begin{equation}
\mathbf{x}^{\prime}\left(  y_{1},y_{2},\cdots,y_{n}\right)  =\int
d\mathbf{x}\Pr\left(  y_{1}|\mathbf{x}\right)  \Pr\left(  y_{2}|\mathbf{x}%
\right)  \cdots\Pr\left(  y_{n}|\mathbf{x}\right)  \,\mathbf{x}%
\end{equation}
The $\left\|  \mathbf{x}-\mathbf{x}^{\prime}\left(  y_{1},y_{2},\cdots
,y_{n}\right)  \right\|  ^{2}$ term may be expanded thus (by adding and
subtracting $\frac{1}{n}\sum_{k=1}^{n}\mathbf{x}_{k}^{\prime}\left(
y_{k}\right)  $)
\begin{equation}
\left\|  \mathbf{x}-\mathbf{x}^{\prime}\left(  y_{1},y_{2},\cdots
,y_{n}\right)  \right\|  ^{2}\equiv\left\|
\begin{array}
[c]{c}%
\left(  \mathbf{x}-\frac{1}{n}\sum_{k=1}^{n}\mathbf{x}_{k}^{\prime}\left(
y_{k}\right)  \right) \\
+\left(  \frac{1}{n}\sum_{k=1}^{n}\mathbf{x}_{k}^{\prime}\left(  y_{k}\right)
-\mathbf{x}^{\prime}\left(  y_{1},y_{2},\cdots,y_{n}\right)  \right)
\end{array}
\right\|  ^{2}%
\end{equation}
Using these two results, together with Bayes' theorem, allows an upper bound
on $D_{VQ}$ to be derived as
\begin{align}
D_{VQ}  & \leq2\int d\mathbf{x}\Pr\left(  \mathbf{x}\right)  \sum_{y_{1}%
=1}^{M_{1}}\sum_{y_{2}=1}^{M_{2}}\cdots\sum_{y_{n}=1}^{M_{n}}\Pr\left(
y_{1}|\mathbf{x},1\right)  \Pr\left(  y_{2}|\mathbf{x},2\right)
\cdots\nonumber\\
& \cdots\Pr\left(  y_{n}|\mathbf{x},n\right)  \left\|  \mathbf{x}-\frac{1}%
{n}\sum_{k=1}^{n}\mathbf{x}_{k}^{\prime}\left(  y_{k}\right)  \right\|  ^{2}%
\end{align}

In this upper bound, the $\Pr\left(  y_{k}|\mathbf{x},k\right)  \,$are used to
produce soft encodings in each of the recognition models ($k=1,2,\cdots,n $),
then a sum $\frac{1}{n}\sum_{k=1}^{n}\mathbf{x}_{k}^{\prime}\left(
y_{k}\right)  $ of the vectors $\mathbf{x}_{k}^{\prime}\left(  y_{k}\right)  $
is used as the reconstruction of the input $\mathbf{x}$. In the special case
where hard encodings are used, so that $\Pr\left(  y_{k}|\mathbf{x},k\right)
=\delta_{y_{k},y_{k}\left(  \mathbf{x}\right)  }$, then the upper bound on
$D_{VQ}$ reduces to $D_{VQ}\leq2\int d\mathbf{x}\Pr\left(  \mathbf{x}\right)
\left\|  \mathbf{x-}\frac{1}{n}\sum_{k=1}^{n}\mathbf{x}_{k}^{\prime}\left(
y_{k}\left(  \mathbf{x}\right)  \right)  \right\|  ^{2}$. Note that the code
vectors used for the encoding operation $y_{k}\left(  \mathbf{x}\right)  $ are
not necessarily the same as the $\mathbf{x}_{k}^{\prime}\left(  y_{k}\right)
$, except in the special case $n=1$.

Suppose that a single recognition model is independently used $n$ times,
rather than $n$ independent recognition models each independently being used
once. This corresponds to constraining the $\mathbf{P}^{l+1}$ vectors and
$\mathbf{P}^{l|l+1}$ matrices to be the same for each of the $n$ recognition
models. The upper bound on $D_{VQ}$ can be manipulated into the form
\begin{align}
D_{VQ}  & \leq\frac{2}{n}\int d\mathbf{x}\Pr\left(  \mathbf{x}\right)
\sum_{y=1}^{M}\Pr\left(  y|\mathbf{x}\right)  \,\left\|  \mathbf{x}%
-\mathbf{x}^{\prime}\left(  y\right)  \right\|  ^{2}\nonumber\\
& +\frac{2\left(  n-1\right)  }{n}\int d\mathbf{x}\Pr\left(  \mathbf{x}%
\right)  \left\|  \mathbf{x}-\sum_{y=1}^{M}\Pr\left(  y|\mathbf{x}\right)
\mathbf{x}^{\prime}\left(  y\right)  \right\|  ^{2}%
\end{align}
where the $k$ index is no longer needed.

\subsection{Full Bayesian Average Over Recognition Models}

\label{Appendix:PMDBayesian}One possible criticism of the recognition model
given in equation \ref{Eq:PosteriorProbabilityPMD} is that it is a mixture of
$K$ recognition models, where each contributing model is assigned the same
weight $\frac{1}{K}$. Normally, a posterior probability $\Pr\left(
y|\mathbf{x}\right)  $ is decomposed as a sum over posterior probabilities
$\Pr\left(  y|\mathbf{x},k\right)  $ derived from each contributing model, as
follows
\begin{equation}
\Pr\left(  y|\mathbf{x}\right)  =\sum_{k=1}^{K}\Pr\left(  y|\mathbf{x}%
,k\right)  \Pr\left(  k|\mathbf{x}\right)
\end{equation}
where each of the $K$ recognition models is assigned a different
data-dependent weight $\Pr\left(  k|\mathbf{x}\right)  $. The conditional
probabilities $\Pr\left(  k|\mathbf{x}\right)  $ and $\Pr\left(
y|\mathbf{x}\right)  $ can be evaluated to yield
\begin{align}
\Pr\left(  k|\mathbf{x}\right)   & =\frac{\sum_{y=1}^{M}\Pr\left(
\mathbf{x}|y,k\right)  \Pr\left(  y|k\right)  \Pr\left(  k\right)  }%
{\sum_{k^{\prime}=1}^{K}\sum_{y^{\prime}=1}^{M}\Pr\left(  \mathbf{x}%
|y^{\prime},k^{\prime}\right)  \Pr\left(  y^{\prime}|k^{\prime}\right)
\Pr\left(  k^{\prime}\right)  }\nonumber\\
\Pr\left(  y|\mathbf{x},k\right)   & =\frac{\Pr\left(  \mathbf{x}|y,k\right)
\Pr\left(  y|k\right)  \Pr\left(  k\right)  }{\sum_{y^{\prime}=1}^{M}%
\Pr\left(  \mathbf{x}|y^{\prime},k\right)  \Pr\left(  y^{\prime}|k\right)
\Pr\left(  k\right)  }%
\end{align}
so that
\begin{equation}
\Pr\left(  y|\mathbf{x}\right)  =\sum_{k=1}^{K}\frac{\Pr\left(  \mathbf{x}%
|y,k\right)  \Pr\left(  y|k\right)  \Pr\left(  k\right)  }{\sum_{k^{\prime}%
=1}^{K}\sum_{y^{\prime}=1}^{M}\Pr\left(  \mathbf{x}|y^{\prime},k^{\prime
}\right)  \Pr\left(  y^{\prime}|k\right)  \Pr\left(  k^{\prime}\right)  }%
\end{equation}
If the replacements $\Pr\left(  k\right)  \rightarrow1$, $\Pr\left(
y|k\right)  \rightarrow A_{k,i_{l+1}}^{l+1}P_{i_{l+1}}^{l+1}$, $\Pr\left(
\mathbf{x}|y,k\right)  \rightarrow P_{i_{l},i_{l+1}}^{l|l+1}$, and $\Pr\left(
y|\mathbf{x}\right)  \rightarrow P_{i_{l+1},i_{l}}^{l+1|l}$ are made, then
$\Pr\left(  y|\mathbf{x}\right)  $ reduces to
\begin{equation}
P_{i_{l+1},i_{l}}^{l+1|l}=\sum_{k=1}^{K}\frac{P_{i_{l},i_{l+1}}^{l|l+1}%
A_{k,i_{l+1}}^{l+1}P_{i_{l+1}}^{l+1}}{\sum_{k^{\prime}=1}^{K}\sum
_{i_{l+1}^{\prime}=1}^{M_{l+1}}P_{i_{l},i_{l+1}^{\prime}}^{l|l+1}A_{k^{\prime
},i_{l+1}^{\prime}}^{l+1}P_{i_{l+1}^{\prime}}^{l+1}}%
\label{Eq:PosteriorProbabilityPMD2}%
\end{equation}
which is not the same as the PMD recognition model in equation
\ref{Eq:PosteriorProbabilityPMD}. The difference between equation
\ref{Eq:PosteriorProbabilityPMD2} and equation
\ref{Eq:PosteriorProbabilityPMD} arises because the full Bayesian approach in
equation \ref{Eq:PosteriorProbabilityPMD2} ensures that the model index $k$
and the input $\mathbf{x}$ are mutually dependent (via the factor $\Pr\left(
k|\mathbf{x}\right)  $), whereas the PMD approach in equation
\ref{Eq:PosteriorProbabilityPMD} ignores such dependencies.

In the full Bayesian approach (see equation \ref{Eq:PosteriorProbabilityPMD2})
the normalisation term in the denominator has a double summation $\sum
_{k=1}^{K}\sum_{i_{l+1}=1}^{M_{l+1}}P_{i_{l},i_{l+1}}^{l|l+1}A_{k,i_{l+1}%
}^{l+1}P_{i_{l+1}}^{l+1} $, which involves all pairs of indices $k$ and
$i_{l+1}$ with $A_{k,i_{l+1}}^{l+1}>0$, which thus corresponds to long-range
lateral interactions in layer $l+1$. On the other hand, in the PMD approach
(see equation \ref{Eq:PosteriorProbabilityPMD}) the normalisation term in the
denominator has only a single summation $\sum_{i_{l+1}=1}^{M_{l+1}}%
P_{i_{l},i_{l+1}}^{l|l+1}A_{k,i_{l+1}}^{l+1}P_{i_{l+1}^{\prime}}^{l+1} $, so
the lateral interactions in layer $l+1$ are determined by the structure of the
matrix $A_{k,i_{l+1}}^{l+1}$, which defines only short-range lateral
connections (i.e. for a given recognition model $k$, only a limited number of
index values $i_{l+1}$ satisfy $A_{k,i_{l+1}}^{l+1}>0$.

\section{Comparison with the Helmholtz Machine}

\label{Appendix:Helmholtz}In this appendix the relationship between two types
of density model is discussed. The first type is a conventional density model
that approximates the input probability density (i.e. the objective function
is $L\left(  \mathbf{P}^{0},\mathbf{Q}^{0}\right)  $), and the second type is
the one introduced here that approximates the joint probability density of a
Markov source (i.e. the objective function is $L\left(  \mathbf{P}%
,\mathbf{Q}\right)  $). In order to relate $L\left(  \mathbf{P}^{0}%
,\mathbf{Q}^{0}\right)  $ to $L\left(  \mathbf{P},\mathbf{Q}\right)  $ it is
necessary to introduce additional layers (i.e. layers $1,2,\cdots,L$) into
$L\left(  \mathbf{P}^{0},\mathbf{Q}^{0}\right)  $ in an appropriate fashion.

The Helmholtz machine (HM) \cite{Dayan1995} does this by replacing $L\left(
\mathbf{P}^{0},\mathbf{Q}^{0}\right)  $ by a different objective function
(which has these additional layers present as hidden variables), and which is
an upper bound on the original objective function $L\left(  \mathbf{P}%
^{0},\mathbf{Q}^{0}\right)  $. It turns out that Helmholtz machine (HM)
objective function $D_{HM}$ and the Markov source objective function $L\left(
\mathbf{P},\mathbf{Q}\right)  $ are closely related. The essential difference
between the two is that $D_{HM}$ does not include the cost of specifying the
state of layers $1,2,\cdots,L$ given that the state of layer 0 is known, which
thus allows it to develop distributed codes (which are expensive to specify)
more easily.

In the conventional density modelling approach to neural networks, there are
two basic classes of model. In the case of both unsupervised and supervised
neural networks the source is $\mathbf{P}^{0}$, which is the network input
(unsupervised case) or the network output (supervised case). Additionally, in
the case of supervised neural networks $\mathbf{P}^{0}$ is conditioned on an
additional network input as $\mathbf{P}^{0|\text{input}}$. Thus in both cases
there is only an external source (i.e. source layers $1,2,\cdots,L$ are not
present), which is modelled by $\mathbf{Q}^{0}$ (unsupervised case) or
$\mathbf{Q}^{0|\text{input}}$ (supervised case). $\mathbf{Q}^{0}$ or
$\mathbf{Q}^{0|\text{input}}$ can be modelled in any way that is convenient.
Frequently a multilayer generative model of the form
\begin{equation}
Q_{i_{0}}^{0}=\sum_{i_{1},i_{2},\cdots,i_{L}}Q_{i_{0},i_{1}}^{0|1}\cdots
Q_{i_{l},i_{l+1}}^{l|l+1}\cdots Q_{i_{L}}^{L}%
\end{equation}
is used, where the $i_{l}$ (for $1\leq l\leq L$) are hidden variables, which
need to be summed over in order to calculate the required marginal probability
$Q_{i_{0}}^{0}$, and the notation is deliberately chosen to be the same as is
used in the Markov chain model
\begin{equation}
Q_{i_{0},i_{1},\cdots,i_{L}}=Q_{i_{0},i_{1}}^{0|1}\cdots Q_{i_{l},i_{l+1}%
}^{l|l+1}\cdots Q_{i_{L}}^{L}%
\end{equation}

Helmholtz machines and Markov sources are related to each other. Thus the
$L\left(  \mathbf{P}^{0},\mathbf{Q}^{0}\right)  $ that is minimised in
conventional density modelling can be manipulated in order to derive $D_{HM}$
\begin{align}
L\left(  \mathbf{P}^{0},\mathbf{Q}^{0}\right)   & \leq L\left(  \mathbf{P}%
^{0},\mathbf{Q}^{0}\right)  +\sum_{i_{0}=1}^{M_{0}}P_{i_{0}}^{0}G_{i_{0}%
}\left(  \mathbf{P}^{1|0},\mathbf{Q}^{1|0}\right) \nonumber\\
& =-\sum_{i_{0}=1}^{M_{0}}P_{i_{0}}^{0}\sum_{i_{1}=1}^{M_{1}}P_{i_{1},i_{0}%
}^{1|0}\log\left(  Q_{i_{0}}^{0}Q_{i_{1},i_{0}}^{1|0}\right) \nonumber\\
& +\sum_{i_{0}=1}^{M_{0}}P_{i_{0}}^{0}\sum_{i_{1}=1}^{M_{1}}P_{i_{1},i_{0}%
}^{1|0}\log P_{i_{1},i_{0}}^{1|0}\nonumber\\
& =L\left(  \left(  \mathbf{P}^{0},\mathbf{P}^{1|0}\right)  ,\left(
\mathbf{Q}^{0},\mathbf{Q}^{1|0}\right)  \right)  -\sum_{i_{0}=1}^{M_{0}%
}P_{i_{0}}^{0}H_{i_{0}}\left(  \mathbf{P}^{1|0}\right) \nonumber\\
& =L\left(  \mathbf{P},\mathbf{Q}\right)  -\sum_{i_{0}=1}^{M_{0}}P_{i_{0}}%
^{0}H_{i_{0}}\left(  \mathbf{P}^{1|0}\right) \nonumber\\
& \equiv D_{HM}%
\end{align}

The inequality $L\left(  \mathbf{P}^{0},\mathbf{Q}^{0}\right)  \leq$ $D_{HM}$
follows from $G_{i_{0}}\left(  \mathbf{P}^{1|0},\mathbf{Q}^{1|0}\right)  \geq0
$ (i.e. the model $\mathbf{Q}^{1|0}$ is imperfect, so that $\mathbf{Q}%
^{1|0}\neq\mathbf{P}^{1|0}$). The inequality $D_{HM}\leq L\left(
\mathbf{P},\mathbf{Q}\right)  $ follows from $H_{i_{0}}\left(  \mathbf{P}%
^{1|0}\right)  \geq0$ (i.e. the source $\mathbf{P}^{1|0}$ is stochastic). If
the model is perfect ($\mathbf{Q}^{1|0}=\mathbf{P}^{1|0}$)\ and the source is
deterministic ($\mathbf{P}^{1|0}$ is such that the state of layer 1 is known
once the state of layer 0 is given), then these two inequalities reduce to
$L\left(  \mathbf{P}^{0},\mathbf{Q}^{0}\right)  =L\left(  \mathbf{P}%
,\mathbf{Q}\right)  $.

The properties of the optimal codes that are used by a Helmholtz machine when
$D_{HM}$ is minimised may be investigated by writing the expression for
$D_{HM}$ as a sum of two terms
\begin{equation}
D_{HM}=\sum_{i_{0}=1}^{M_{0}}P_{i_{0}}^{0}K_{i_{0}}\left(  \mathbf{P}%
^{1|0},\mathbf{Q}^{0|1}\right)  +\sum_{i_{0}=1}^{M_{0}}P_{i_{0}}^{0}G_{i_{0}%
}\left(  \mathbf{P}^{1|0},\mathbf{Q}^{1}\right)
\end{equation}
The $\sum_{i_{0}=1}^{M_{0}}P_{i_{0}}^{0}K\left(  \mathbf{P}^{1|0}%
,\mathbf{Q}^{0|1}\right)  $ part and the $\sum_{i_{0}=1}^{M_{0}}P_{i_{0}}%
^{0}G\left(  \mathbf{P}^{1|0},\mathbf{Q}^{1}\right)  $ part compete with each
other when $D_{HM}$ is minimised. Assuming that $P_{i_{0}}^{0}>0$, the
$\sum_{i_{0}=1}^{M_{0}}P_{i_{0}}^{0}G\left(  \mathbf{P}^{1|0},\mathbf{Q}%
^{1}\right)  $ part likes to make $\mathbf{Q}^{1}$ approximate $\mathbf{P}%
^{1|0}$, which tends to make $\mathbf{P}^{1|0}$ behave like a distributed
encoder. On the other hand, assuming that $P_{i_{1}}^{1}>0$, the $\sum
_{i_{0}=1}^{M_{0}}P_{i_{0}}^{0}K\left(  \mathbf{P}^{1|0},\mathbf{Q}%
^{0|1}\right)  $ part likes to make $\mathbf{Q}^{0|1}$ approximate
$\mathbf{P}^{0|1}$, which tends to make $\mathbf{P}^{1|0}$ behave like a
sparse encoder. The tension between these two terms is optimally balanced when
$D_{HM}$ is minimised.

The properties of the optimal codes that are used in the Markov source
approach when $L\left(  \mathbf{P},\mathbf{Q}\right)  $ (see equation
\ref{Eq:ObjectiveMarkovSource}) is minimised are different. The 2-layer
expression for $L\left(  \mathbf{P},\mathbf{Q}\right)  $ is
\begin{equation}
L\left(  \mathbf{P},\mathbf{Q}\right)  =\sum_{i_{0}=1}^{M_{0}}P_{i_{0}}%
^{0}K_{i_{0}}\left(  \mathbf{P}^{1|0},\mathbf{Q}^{0|1}\right)  +L\left(
\mathbf{P}^{L},\mathbf{Q}^{L}\right)
\end{equation}
which contains the same sparse encoder term $\sum_{i_{0}=1}^{M_{0}}P_{i_{0}%
}^{0}K_{i_{0}}\left(  \mathbf{P}^{1|0},\mathbf{Q}^{0|1}\right)  $ as $D_{HM}$.
However, the distributed encoder term $\sum_{i_{0}=1}^{M_{0}}P_{i_{0}}%
^{0}G_{i_{0}}\left(  \mathbf{P}^{1|0},\mathbf{Q}^{1}\right)  $ is missing, and
is replaced by $L\left(  \mathbf{P}^{L},\mathbf{Q}^{L}\right)  $ which does
not have the effect of encouraging any particular type of code (other than one
in which $\mathbf{P}^{L}$ approximates $\mathbf{Q}^{L}$).

These differences between $D_{HM}$ and $L\left(  \mathbf{P},\mathbf{Q}\right)
$ show how the Markov source approach encourages sparse codes to develop,
whereas the Helmholtz machine does not. It is not clear whether using $D_{HM}
$ is the best approach to forming distributed codes, because there are other
ways of encouraging distributed codes to develop, such as the factorial
encoder discussed in section \ref{Sect:PMD}, which is based on $L\left(
\mathbf{P},\mathbf{Q}\right)  $ rather than $D_{HM}$.

\end{document}